\documentclass{article}

\usepackage[main, preprint]{neurips_2026}

\usepackage[utf8]{inputenc} %
\usepackage[T1]{fontenc}    %
\usepackage{hyperref}       %
\usepackage{url}            %
\usepackage{booktabs}       %
\usepackage{amsfonts}       %
\usepackage{nicefrac}       %
\usepackage{microtype}      %
\usepackage[table]{xcolor}  %
\usepackage{standalone}
\usepackage[utf8]{inputenc}
\usepackage[T1]{fontenc}

\usepackage{latexsym}
\usepackage{amsmath}
\usepackage{amssymb}
\usepackage{amsthm}
\usepackage{natbib}   
\usepackage{graphicx}
\usepackage{subcaption}
\usepackage{array}
\usepackage{tabu}
\usepackage{makecell}
\usepackage{paralist}
\usepackage{cases}
\usepackage{diagbox}
\usepackage{enumitem}
\usepackage{soul}
\usepackage{multirow}
\usepackage{verbatim}
\usepackage{tabulary}
\usepackage{booktabs}
\usepackage{tabularx}
\usepackage[mathscr]{euscript}
\usepackage{mathtools}
\usepackage{algorithm}
\usepackage{algpseudocode}
\usepackage{stmaryrd}
\usepackage{tikz-dependency}
\usetikzlibrary{automata,decorations.markings,arrows,positioning,matrix,calc,patterns,angles,quotes,calc}
\usepackage{adjustbox}
\usepackage{tabularx}
\usepackage{xspace}
\usepackage{tabulary}
\usepackage{afterpage}
\usepackage{bm}
\usepackage{color}
\usepackage{graphicx}
\usepackage{slashbox}
\usepackage[toc,page]{appendix}
\usepackage{makecell}
\usepackage{boldline}
\usepackage[shortcuts]{extdash}  %

\usepackage{blindtext}
\usepackage{graphicx}
\usepackage{capt-of}
\usepackage{booktabs}
\usepackage{varwidth}
\usepackage{pifont}%
\usepackage{wrapfig}

\usepackage{upquote}
\usepackage{listings}
\usepackage[normalem]{ulem}
\useunder{\uline}{\ul}{}
\usepackage{ragged2e} %

\usepackage{comment} %

\usepackage{needspace}  %

\usepackage[most]{tcolorbox}

\usepackage{threeparttable}

\definecolor{orange}{rgb}{1,0.5,0}
\definecolor{mdgreen}{rgb}{0.05,0.6,0.05}
\definecolor{mdblue}{rgb}{0,0,0.7}
\definecolor{dkblue}{rgb}{0,0,0.5}
\definecolor{dkgray}{rgb}{0.3,0.3,0.3}
\definecolor{slate}{rgb}{0.25,0.25,0.4}
\definecolor{gray}{rgb}{0.5,0.5,0.5}
\definecolor{ltgray}{rgb}{0.7,0.7,0.7}
\definecolor{purple}{rgb}{0.7,0,1.0}
\definecolor{lavender}{rgb}{0.65,0.55,1.0}

\definecolor{mypurple}{RGB}{111,61,121}
\definecolor{myblue}{RGB}{46,88,180}
\definecolor{myred}{RGB}{181,68,106}
\definecolor{myyellow}{RGB}{204,143,55}

\newcommand{\ensuretext}[1]{#1}
\newcommand{\marker}[2]{\ensuremath{^{\textsc{#1}}_{\textsc{#2}}}}
\newcommand{\draftcomment}[3]{\ensuretext{\textcolor{#3}{[#1 #2]}}}
\renewcommand{\draftcomment}[3]{}  %

\newcommand{\cmark}{\ding{51}}%
\newcommand{\xmark}{\ding{55}}%

\newcommand{\cmarkgreen}{\textcolor[HTML]{009901}{\cmark}}%
\newcommand{\xmarkred}{\textcolor[HTML]{9A0000}{\xmark}}%
\newcommand{\cmarkgreenbf}{\textcolor[HTML]{009901}{\ding{52}}}%

\newif\ifqiruncomments
\qiruncommentstrue    %

\DeclareSymbolFont{extraup}{U}{zavm}{m}{n}
\DeclareMathSymbol{\vardiamond}{\mathalpha}{extraup}{87}

\newcolumntype{L}[1]{>{\raggedright\let\newline\\\arraybackslash\hspace{0pt}}m{#1}}
\newcolumntype{C}[1]{>{\centering\let\newline\\\arraybackslash\hspace{0pt}}m{#1}}
\newcolumntype{R}[1]{>{\raggedleft\let\newline\\\arraybackslash\hspace{0pt}}m{#1}}

\theoremstyle{definition}

\theoremstyle{remark}

\algrenewcommand{\algorithmiccomment}[1]{\leavevmode$\triangleright$ #1}

\setul{1pt}{.4pt}

\DeclareFixedFont{\ttb}{T1}{txtt}{bx}{n}{12} %
\DeclareFixedFont{\ttm}{T1}{txtt}{m}{n}{12}  %

\newcommand{\para}[1]{\noindent\textbf{#1}}

\setlist[enumerate]{leftmargin=*, itemsep=1mm, parsep=1mm, topsep=1mm, partopsep=1mm}
\setlist[itemize]{leftmargin=*, itemsep=1mm, parsep=1mm, topsep=1mm, partopsep=1mm}

\colorlet{punct}{red!60!black}
\definecolor{background}{HTML}{EEEEEE}
\definecolor{delim}{RGB}{20,105,176}
\colorlet{numb}{magenta!60!black}

\lstdefinelanguage{json}{
    basicstyle=\normalfont\ttfamily,
    numbers=left,
    numberstyle=\scriptsize,
    stepnumber=1,
    numbersep=8pt,
    showstringspaces=false,
    breaklines=true,
    frame=lines,
    backgroundcolor=\color{background},
    literate=
     *{0}{{{\color{numb}0}}}{1}
      {1}{{{\color{numb}1}}}{1}
      {2}{{{\color{numb}2}}}{1}
      {3}{{{\color{numb}3}}}{1}
      {4}{{{\color{numb}4}}}{1}
      {5}{{{\color{numb}5}}}{1}
      {6}{{{\color{numb}6}}}{1}
      {7}{{{\color{numb}7}}}{1}
      {8}{{{\color{numb}8}}}{1}
      {9}{{{\color{numb}9}}}{1}
      {:}{{{\color{punct}{:}}}}{1}
      {,}{{{\color{punct}{,}}}}{1}
      {\{}{{{\color{delim}{\{}}}}{1}
      {\}}{{{\color{delim}{\}}}}}{1}
      {[}{{{\color{delim}{[}}}}{1}
      {]}{{{\color{delim}{]}}}}{1},
}

\newcommand{\ourdata}{\textsc{UniCo}\xspace}
\newcommand{\name}{\textsc{UniCo}\xspace}

\title{Towards a Universal Causal Reasoner}

\author{%
  Qirun Dai\,$^{1*\dagger}$ \quad
  Xiao Liu\,$^{1*\dagger}$ \quad
  Jiawei Zhang\,$^{1*}$ \quad
  Dylan Zhang\,$^{2}$ \\
  \bf
  Hao Peng\,$^{2}$ \quad
  Chenhao Tan\,$^{1}$ \\
  \normalfont\rule{0mm}{4mm}%
  $^{1}$The University of Chicago \quad
  $^{2}$University of Illinois Urbana-Champaign
}

\begin{document}

\maketitle
\begingroup
  \renewcommand{\thefootnote}{\fnsymbol{footnote}}%
  \NoHyper
  \setcounter{footnote}{1}\footnotetext{Equal contribution.}%
  \setcounter{footnote}{2}\footnotetext{Project co-lead. Correspondence to: 
  \texttt{\{qirundai,liuxiao,chenhao\}@uchicago.edu}.}%
  \endNoHyper
\endgroup
\setcounter{footnote}{0}

\begin{abstract}

Despite the importance of causal reasoning, training LLMs to reason causally remains underexplored. %
Existing data efforts mostly
focus on benchmarking LLMs on specific aspects of causality, 
making them less suitable for training generalizable causal reasoners.
To address this, we propose \name, a data generation framework that both (1) addresses 18 causal query types across Pearl's Causal Ladder and (2) translates natively symbolic examples into code and natural language forms 
to simulate real-world use cases where causal terms are not explicitly specified. 
To ensure data quality, \name 
grounds answers with exact causal inference and filters cases with
reasoning shortcuts.
Upon supervised finetuning with 66.6K \name-generated instances, Qwen3-4B, Qwen3-8B and Olmo-3-7B-Instruct 
achieve an average of 22.9\% improvements across all 18 in-distribution query types, and 8.1\% over state-of-the-art causal data generation frameworks on 7 established causal benchmarks outside the training distribution.
More importantly, in real-world medical understanding, legal decision, and tabular reasoning, \name-trained models consistently display more faithful reasoning traces, outperforming the base models by an average of 20.2\% in faithfulness metrics. %
These suggest that causality-centered training not only strengthens causal reasoning, but also equips LLMs with a causal mindset
in general reasoning tasks.\footnote{All \name-generated datasets are available at
\href{https://huggingface.co/collections/ChicagoHAI/unico}{ChicagoHAI/UniCo}.
Code will be soon available.}

\end{abstract}

\vspace{-3mm}

\begin{figure}[H]
    \centering
    \begin{minipage}[c]{0.55\linewidth}
        \centering
        \includegraphics[width=\linewidth,keepaspectratio]{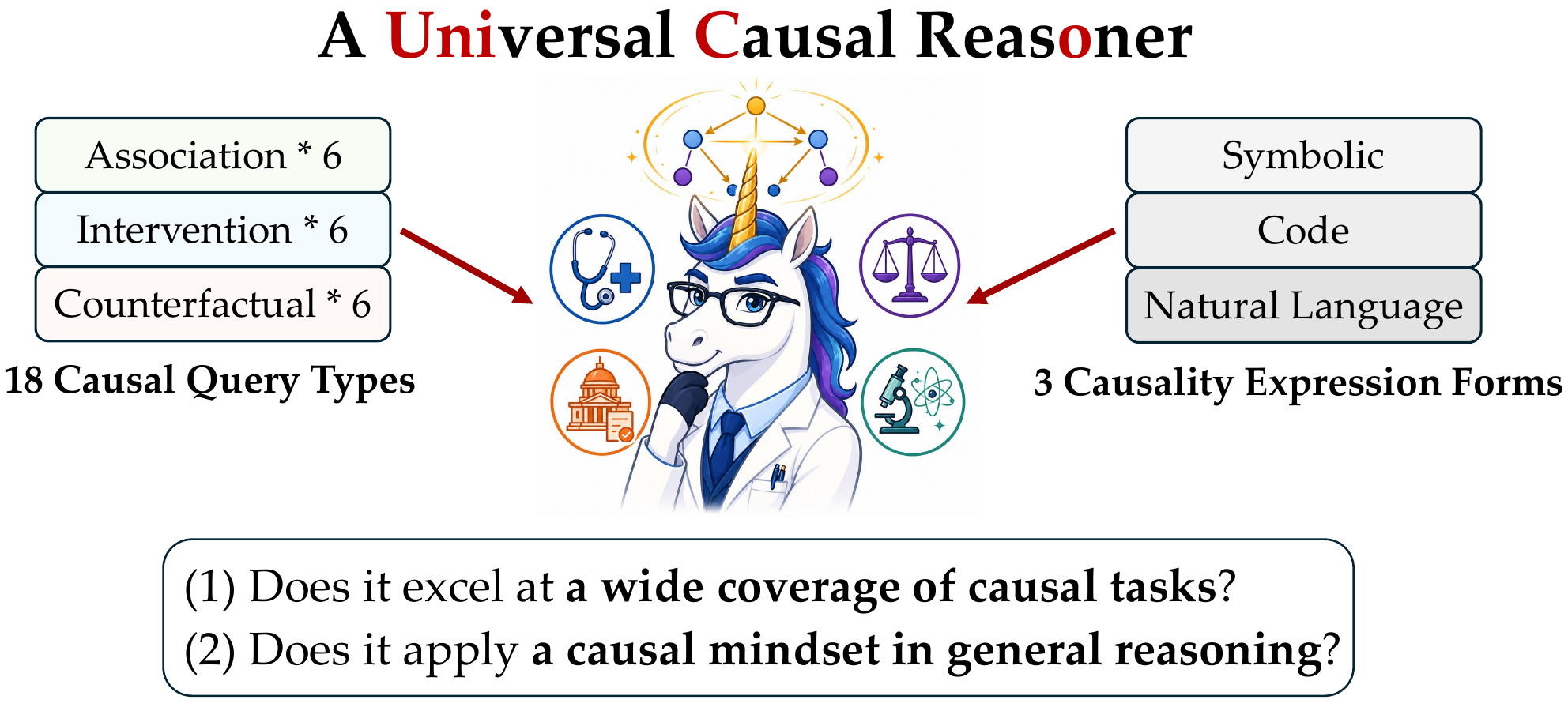}
    \end{minipage}\hspace{0.01\linewidth}%
    \begin{minipage}[c]{0.43\linewidth}
        \centering
        \includegraphics[width=\linewidth,keepaspectratio]{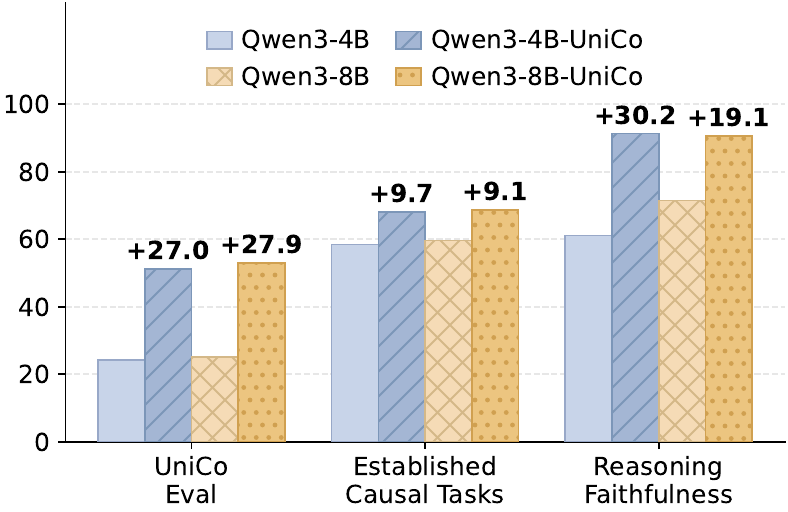}
    \end{minipage}
    \caption{\textbf{Left:} Overview of \name's technical components and generalization goals. \textbf{Right:} Performance gains from SFT on \name-generated data, spanning in-distribution query types, established OOD causal tasks, and faithfulness in real-world general reasoning.}
    \label{fig:unico-overview}
\end{figure}

\vspace{-3mm}

\section{Introduction}
Causal reasoning is a core component of human cognition,
enabling us to understand how the world works by tracing outcomes to their origins, and thus make rational decisions. 
Despite rapid progress in large language models (LLMs), their ability to accurately solve causal problems remains limited. They tend to generate fallacies such as confusing correlation with causation~\citep{liu2024llms,du2025ice}, and their reasoning traces are frequently unfaithful, with explanations that do not reflect the true origin of their predictions~\citep{arcuschin2025chain,chen2025reasoning}.

Motivated by the limitations of current models, 
we aim to build a universal causal reasoner. 
We define universality along two complementary dimensions. First, \textbf{within causality}, a model should generalize across a wide coverage of causal tasks, rather than overfitting to specific query types or representation forms. %
Second, \textbf{beyond causality}, it should autonomously adopt a causal mindset that promotes more faithful reasoning, producing answers consistent with their reasoning processes.

It remains challenging to achieve this goal, with \textbf{data} as a key bottleneck. Existing efforts either focus on benchmarking specific aspects of causal reasoning without providing training data~\citep{jin2024can,xiong2025com2}, or only construct synthetic datasets with limited diversity~\citep{vashishtha2026executable,dong2025care}. In addition, several causal datasets suffer from quality issues, such as missing necessary information or ambiguous question formulations, which further hinders effective training.
The most relevant work, CauGym~\citep{chen2026can}, introduces a dedicated training dataset for causal reasoning. However, its coverage remains limited since its construction pipeline closely mirrors previous benchmarks like CaLM~\citep{chen2024causal}. As a result, models trained on such datasets tend to overfit to these benchmarks, eventually showing limited generalizability.

To address this gap, we introduce \name, 
a principled data generation framework that emphasizes both diversity and quality~\citep{liu2024makes, albalak2024a, dai-etal-2025-improving}.
Diversity-wise, \name addresses 18 query types across all three levels--association, intervention, and counterfactual--of Pearl's Causal Ladder~\citep{Pearl_2009}, which teaches models the comprehensive principles of causal reasoning.
It also presents each question in three complementary forms: symbolic notations, executable code, and natural language narratives under real-world contexts,
teaching models to autonomously recognize causal structures and perform causal reasoning across varying task expressions in general scenarios. %
Quality-wise, \name ensures that each question provides all necessary conditions under an unambiguous formulation, and computes all ground truth answers with exact causal inference. Moreover, \name promotes dataset difficulty by explicitly controlling the ratio of ``causally naive questions'' that can be solved by a degraded lower-level calculation on the causal ladder (e.g., average treatment effect that can be computed by directly taking the difference of conditional probabilities), thus preventing models from learning reasoning shortcuts.

Empirically, 
finetuning
Qwen3-4B, Qwen3-8B, and Olmo-3-7B-Instruct on 66,603 \name-generated examples achieves an average of 22.9\% performance gain across all 18 in-distribution query types.
\name also outperforms both CauGym and CDCR~\citep{li2026mitigating}
by an average of 8.1\% on 7 established causal benchmarks outside the training distribution, including CaLM~\citep{chen2024causal}, CLadder~\citep{jin2023cladder} and Corr2Cause~\citep{jin2024can}.
Notably, we find diversity across both the query types and representation forms plays a critical role in building generalizable causal experts.

Beyond causal tasks, we further find that
\name-trained models consistently display more faithful reasoning in read-world medical understanding, legal decision, and tabular reasoning measured by the RFEval benchmark~\citep{hanrfeval}, with an average gain of 20.2\% over the base models.
This validates that exposure to diverse and quality causal reasoning data not only generalizes within causality, but also to more causally consistent reasoning behaviors in general scenarios.

Overall, our contributions include: 
(1) \name, a causal data generation framework with first-of-its-kind diversity and quality.
(2) A universal causal reasoner that not only excels at a wide range of causal tasks, but also performs faithful and causally consistent reasoning generally. %
(3) Showing for the first time that causality-centered training not only strengthens causal reasoning, but also equips LLMs with a causal mindset in general reasoning tasks.

\section{Related Work}

\begin{table*}[ht]
    \vspace{-3mm}
    \begin{threeparttable}
    \caption{
    Comparison of representative causal reasoning datasets.\tnote{1}
    ``Suff. Cond.'' indicates whether each instance contains sufficient conditions for the question to be solvable, and ``Unamb.'' indicates whether the query is semantically unambiguous. 
    ``Rep. Form'' covers three major representation forms of a causal question (symbolic, code and natural language), and ``Query Type'' counts 
    the number of query types covered 
    across the causal ladder (association, intervention, and counterfactual).
    The final ``Size'' considers both training and evaluation splits.
    }
    \label{table-data-compare}
    \centering
    \small
    \setlength{\tabcolsep}{2pt}
    \begin{tabular}{@{}lccccccccccc@{}}
        \toprule
        \multicolumn{1}{c}{\multirow{2}{*}{\textbf{Dataset}}} &
          \multirow{2}{*}{\textbf{Training}} &
          \multirow{2}{*}{\textbf{Scalable}} &
          \multirow{2}{*}{\textbf{\begin{tabular}[c]{@{}c@{}}Suff.\\ Cond.\end{tabular}}} &
          \multirow{2}{*}{\textbf{Unamb.}} &
          \multicolumn{3}{c}{\textbf{Rep. Form}} &
          \multicolumn{3}{c}{\textbf{Query Type}} &
          \multirow{2}{*}{\textbf{Size}} \\
        \cmidrule(lr){6-8} \cmidrule(lr){9-11}
         & & & & &
          \textbf{Sym.} & \textbf{Code} & \textbf{NL} &
          \textbf{Assn.} & \textbf{Int.} & \textbf{CF} & \\ \midrule
        BBEH~{\scriptsize\citep{kazemi2025big}} &
          \xmarkred &
          \xmarkred &
          \cmarkgreen &
          \cmarkgreen &
          \xmarkred &
          \xmarkred &
          \cmarkgreen &
          0 &
          1 &
          1 &
          200 \\
        CounterBench~{\scriptsize\citep{chen2025counterbench}} &
          \xmarkred &
          \cmarkgreen &
          \cmarkgreen &
          \xmarkred &
          \cmarkgreen &
          \xmarkred &
          \xmarkred &
          0 &
          0 &
          4 &
          1,200 \\
        CLadder~{\scriptsize\citep{jin2023cladder}} &
          \xmarkred &
          \cmarkgreen &
          \xmarkred &
          \xmarkred &
          \cmarkgreen &
          \xmarkred &
          \cmarkgreen &
          3 &
          3 &
          4 &
          10,112 \\
        CaLM~{\scriptsize\citep{chen2024causal}} &
          \xmarkred &
          \cmarkgreen &
          \xmarkred &
          \xmarkred &
          \cmarkgreen &
          \xmarkred &
          \cmarkgreen &
          0 &
          \textbf{6} &
          5 &
          6,200 \\
        ExecCF~{\scriptsize\citep{vashishtha2026executable}} &
          \cmarkgreen &
          \xmarkred &
          \xmarkred &
          \cmarkgreen &
          \xmarkred &
          \cmarkgreen &
          \cmarkgreen &
          0 &
          1 &
          1 &
          9,290 \\
        CauGym~{\scriptsize\citep{chen2026can}} &
          \cmarkgreen &
          \cmarkgreen &
          \xmarkred &
          \xmarkred &
          \cmarkgreen &
          \xmarkred &
          \cmarkgreen &
          0 &
          2 &
          5 &
          7,000 \\
        CDCR~{\scriptsize\citep{li2026mitigating}} &
          \cmarkgreen &
          \xmarkred &
          \xmarkred &
          \xmarkred &
          \cmarkgreen &
          \xmarkred &
          \cmarkgreen &
          3 &
          3 &
          4 &
          25,368 \\
        \textbf{\name (Ours)} &
          \cmarkgreenbf &
          \cmarkgreenbf &
          \cmarkgreenbf &
          \cmarkgreenbf &
          \cmarkgreenbf &
          \cmarkgreenbf &
          \cmarkgreenbf &
          \textbf{6} &
          \textbf{6} &
          \textbf{6} &
          \textbf{79,924} \\ \bottomrule
    \end{tabular}

    \begin{tablenotes}
        \footnotesize
        \item[1] For CaLM, we only consider self-constructed and publicly available subsets. BBEH and CDCR are not scalable because they are completely adapted from existing benchmarks, while ExecCF loses its scalability due to reliance on human-designed templates.
    \end{tablenotes}
        
    \end{threeparttable}
\end{table*}

\para{Causal reasoning benchmarks.}
Existing benchmarks on causal reasoning mainly fall into two lines. 
One focuses on \emph{commonsense} causal reasoning in natural language forms, where models judge cause and effect in realistic scenarios using implicit world knowledge~\citep{chi2024unveiling,xiong2025com2,kazemi2025big}. 
The other studies more \emph{formalized} causal reasoning, where the causal structure and inference rules are explicitly presented, including symbolic forms~\citep{jin2024can,chen2025counterbench} and mathematical formulations~\citep{jin2023cladder,vashishtha2026executable,chen2024causal}. 

However, it is often suboptimal to directly adapt these benchmarks into training data. 
As shown in Table~\ref{table-data-compare}, they mostly have a narrow, unbalanced coverage of causal query types and representation forms of the causal context. 
Even for datasets with broader coverage, such as~\citet{chen2024causal,jin2023cladder}, they still contain an unignorable proportion of instances with insufficient causal conditions or ambiguous question specifications, for which we perform a detailed analysis in Appendix~\ref{appendix-data-quality}. 
These limitations motivate the need for a training-oriented data generation pipeline with both broader coverage and stronger quality control~\citep{liu2024makes,albalak2024a, dai-etal-2025-improving}.

\para{Efforts to train causal reasoners.}
Recent efforts have started to move beyond evaluation and constructed training data for causal reasoning. 
CauGym~\citet{chen2026can} and CDCR~\citep{li2026mitigating} both synthesize training datasets and show that post-training improves LLMs' causal reasoning performance. 
However, their data frameworks are largely derived from existing benchmarks: CauGym adopts the same data generation pipeline as the seven probabilistic query types in CaLM~\citep{chen2024causal}, while CDCR directly constructs its data using part of the examples in CLadder~\citep{jin2023cladder}.
Therefore, they inherit the same limitations in coverage and quality from existing benchmarks, and models trained on them tend to overfit to the source benchmarks and exhibit limited generalization to a broader scope of causal reasoning.

Other works introduce causal training signals for specific downstream purposes~\citep{zhou2023causal,liu2025eliciting}. For example, \citet{cheng2025mitigating} mitigate spurious correlations in sentiment classification by identifying and emphasizing causality-related words during training, and C2PO~\citep{feng2025c2po} constructs causally contrastive preference data to reduce bias shortcuts in LLMs. These works show the potential of introducing causality into LLM training, but the resulting models remain tied to their specific target tasks. These further highlight the importance of building a universal causal reasoner 
with transferable causal reasoning skills both within and beyond causal tasks.

\section{\name: A Data Framework Towards Universal Causal Reasoners}

We design \name on the basis of structural causal models (SCMs)~\citep{Pearl_2009}.
As shown in Figure~\ref{fig-example-domains}, given an SCM consisting of a directed acyclic graph (DAG) and a set of probability conditions about its variables, \name creates symbolic questions by applying a causal query to the SCM, and further renders the same questions in code and natural language forms. %
The main features of \name include
the diversity of causal queries (\S\ref{sec:query-sampling}), translation of SCMs across representation forms (\S\ref{sec:domain-translation}), and the quality control mechanism 
that ensures causally meaningful questions
(\S\ref{sec:quality-control}). %
Please refer to Appendix~\ref{appendix-causal-knowledge} for preliminaries of the SCM framework and other causal background knowledge.

\subsection{Structural Causal Model (SCM) Sampling}\label{sec:scm-sampling}

\para{Graph sampling.}
We build upon the iterative graph sampling algorithm in \citet{lu2026generalization} and \citet{NEURIPS2023_045c87de}, which starts from a random set of root nodes, and introduces the remaining nodes in a topological order, 
with each new node choosing 1 or 2 parents uniformly from those already present. 
This approach 
keeps local mechanisms tractable for exact causal inference while still allows multi-step complex causal structures.
To promote topological diversity, we reject graphs isomorphic to previously sampled ones.
Eventually, we sample 4,238 and 372 unique graphs for all our training and evaluation data respectively, with the number of nodes ranging from 3 to 10.

\para{Probabilistic sampling.}
Given a causal graph, we then sample its probabilistic mechanisms in the form of conditional probability tables. 
For simplicity, we only consider SCMs where each variable takes binary values.
For each node \(v\) and each of its parent assignment \(\bf{a} \in \{0,1\}^{|\mathrm{pa}(v)|}\), we define the value for $P(v = 1 \mid \mathrm{pa}(v) = \bf{a})$.
Thus a node with \(k \in \{1,2\}\) parents contributes \(2^k\) probability conditions, as illustrated in Figure~\ref{fig-example-symbolic}.
Notably, for counterfactual query types, we assume an additional group of exogenous noise variables for each node in the causal graph, to enable the canonical abduction-intervention-prediction~\citep{vashishtha2026executable} workflow that prevents them from downgrading to intervention queries.

\subsection{Query Sampling and Symbolic Data Generation}\label{sec:query-sampling}

\begin{figure}[t]
    \centering
    \begin{subfigure}[b]{0.32\linewidth}
        \includegraphics[width=\linewidth]{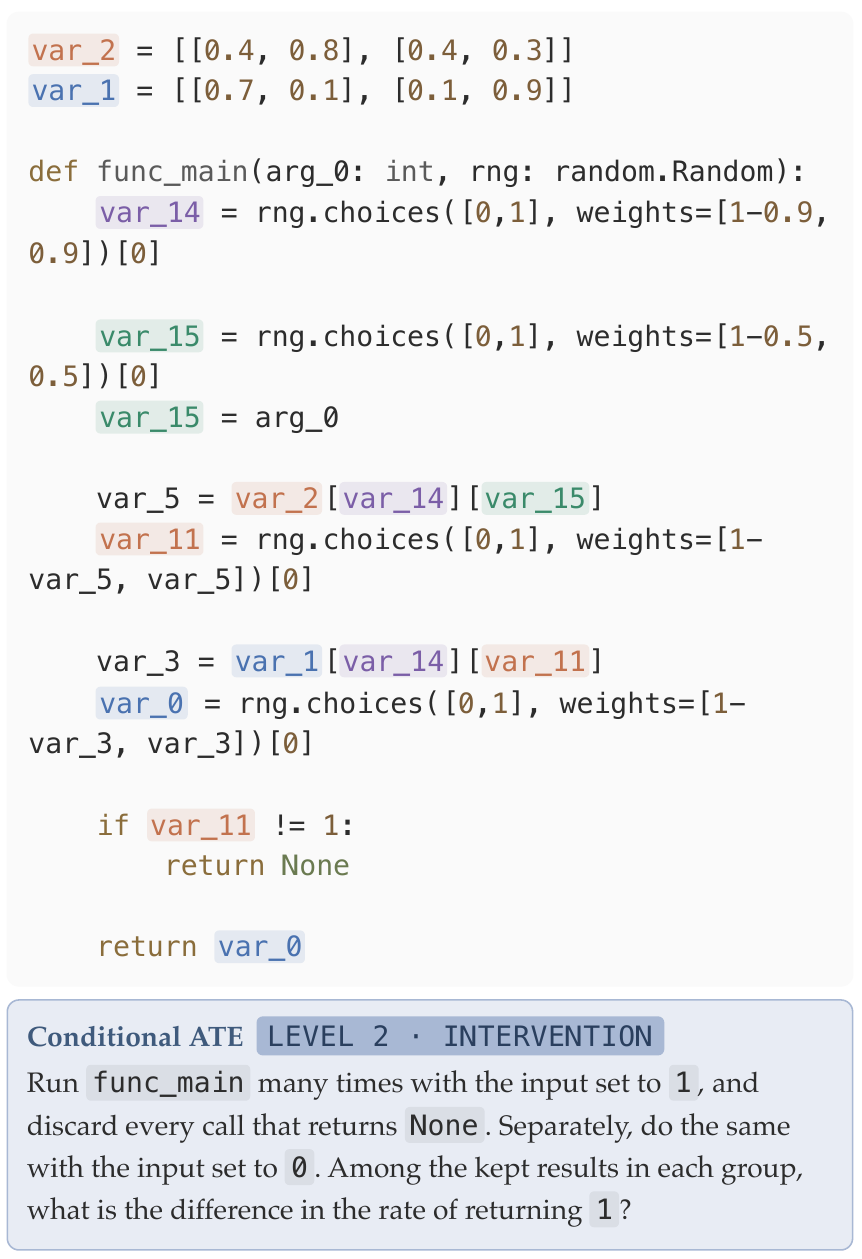}
        \caption{Code}
        \label{fig-example-code}
    \end{subfigure}
    \hfill
    \begin{subfigure}[b]{0.32\linewidth}
        \includegraphics[width=\linewidth]{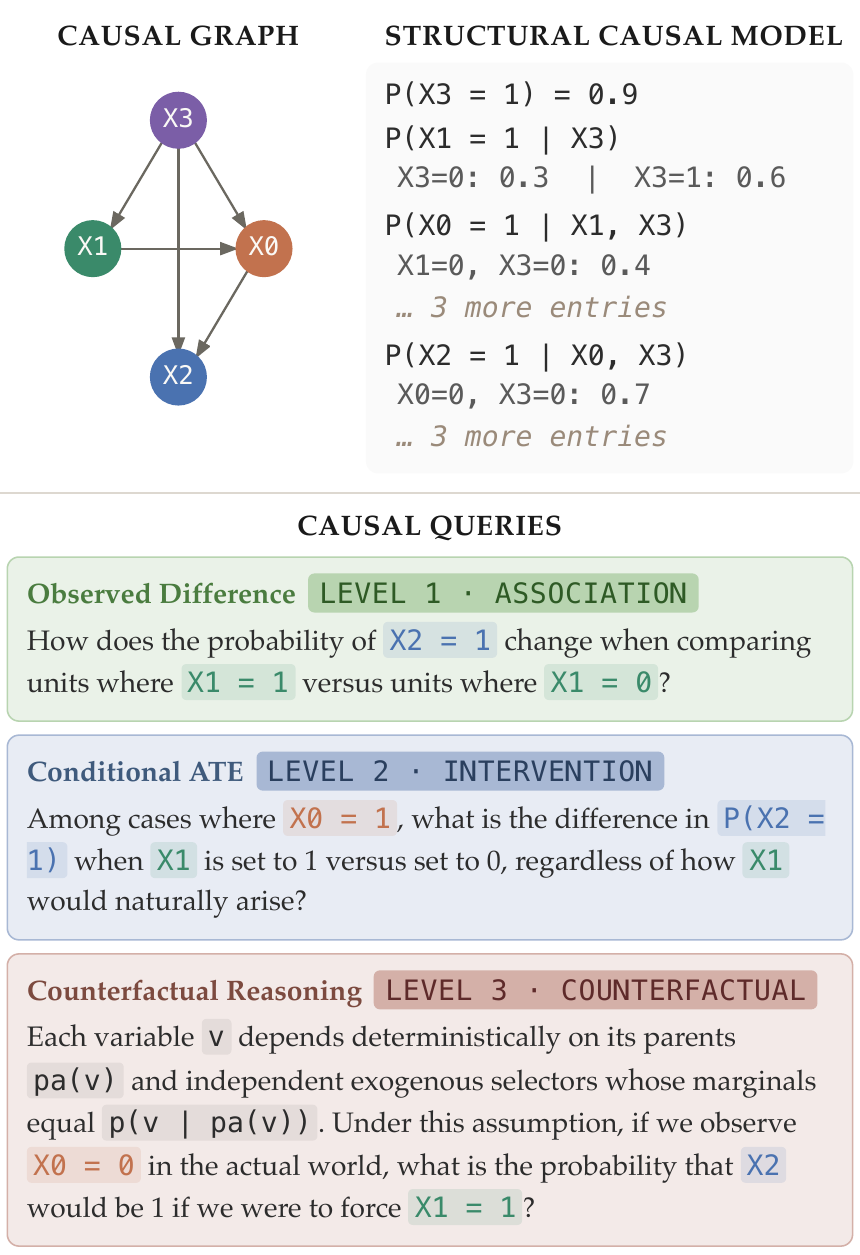}
        \caption{Symbolic}
        \label{fig-example-symbolic}
    \end{subfigure}
    \hfill
    \begin{subfigure}[b]{0.32\linewidth}
        \includegraphics[width=\linewidth]{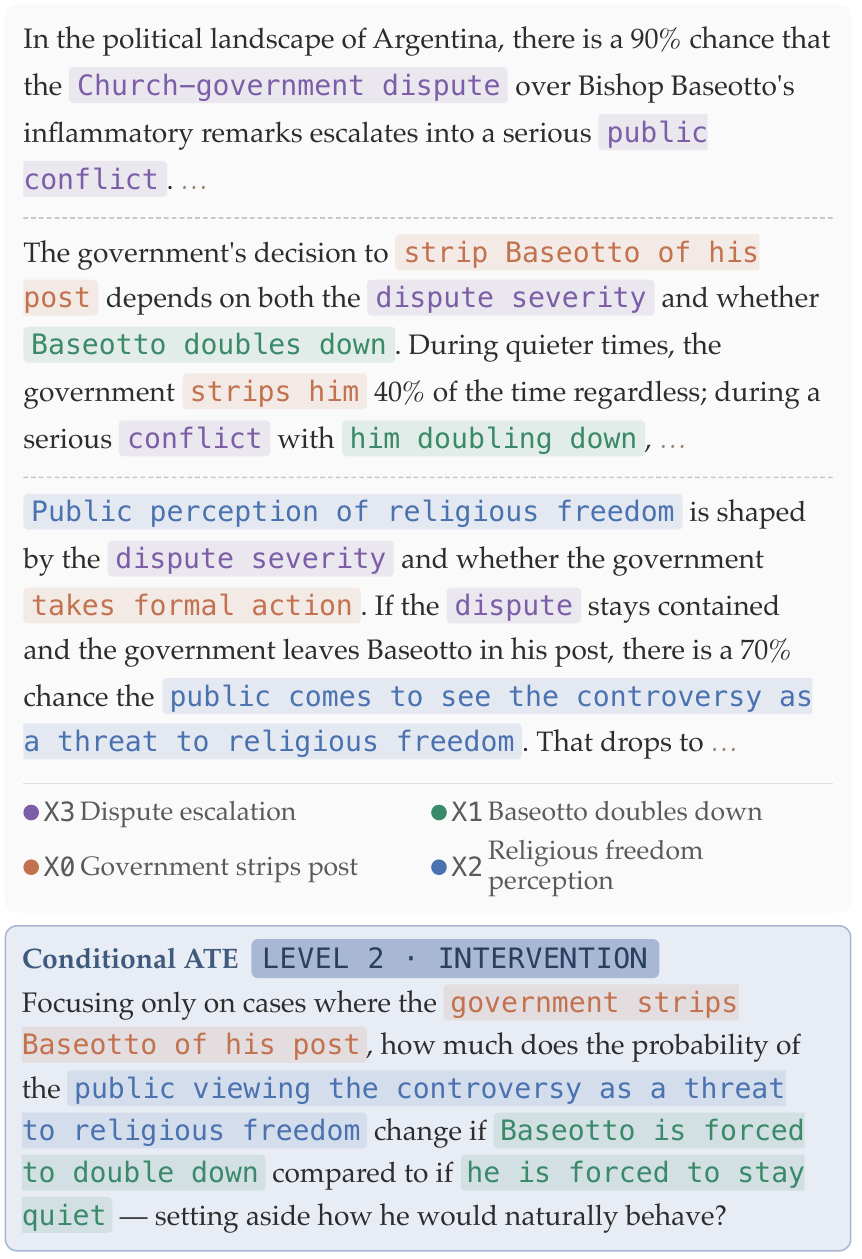}
        \caption{Natural Language}
        \label{fig-example-nl}
    \end{subfigure}
    \caption{
    Examples illustrating the three representation forms. 
    \name starts from the top half of (b) by sampling a causal graph and its parent-child probabilistic conditions (\S\ref{sec:scm-sampling}) to form a structural causal model. 
    On top of it, \name samples from 18 query types spanning all three levels of the Causal Ladder, and formulates symbolic questions, as shown in the bottom half of (b) (\S\ref{sec:query-sampling}).
    Eventually, each symbolic question can be translated into either executable code (a) or natural language (c) forms (\S\ref{sec:domain-translation}), with the underlying causal semantics intact. See Appendix~\ref{appendix-complete-examples} for full examples. %
    }
    \label{fig-example-domains}
    \vspace{-3mm}
\end{figure}

Given an SCM, \name then samples the causal query of interest and generates symbolic questions.
\name covers 18 causal query types incorporating the three levels of Judea Pearl's Causal Ladder~\citep{Pearl_2009,pearl2018book}---association, intervention, and counterfactual. Among them, 5 are graph-only types (\textit{italicized} below) that focus on binary causal judgment, and the remaining 13 target probabilistic causal inference (see Appendix~\ref{appendix-query-types} for detailed illustrations of query types). 

\begin{itemize}[noitemsep,topsep=0pt,parsep=0pt,partopsep=0pt]
    \item \textbf{Association}: Marginal Probability (MP), Conditional Probability (CP), Joint Probability (JP), Observed Difference (OD), \textit{Independence Test (IT)}, \textit{Markov Blanket (MB)}
    \item \textbf{Intervention}: Average Treatment Effect (ATE), Conditional ATE (CTE), Joint ATE (JTE), \textit{Identifiability (ID)}, \textit{Frontdoor Adjustment (FD)}, \textit{Backdoor Adjustment (BD)}
    \item \textbf{Counterfactual}: Counterfactual Probability (CF), Average Treatment effect on the Treated (ATT), Natural Indirect Effect (NIE), Natural Direct Effect (NDE), Probability of Necessity (PN), Probability of Sufficiency (PS)
\end{itemize}

We ask questions of different query types by sampling a group of \textbf{operation nodes} and assigning different \textbf{roles} to them.
Take Figure~\ref{fig-example-symbolic} as a running example. 
If we sample $X_2$ as the \textit{outcome} variable and $X_1$ as the observed \textit{evidence}, then an Observed Difference (level 1) query is assembled with the following symbolic expression: $P(X_2=1 \mid X_1 = 1) - P(X_2=1 \mid X_1=0)$.
If we keep $X_2$ as the outcome, but regard $X_1$ as an intervened \textit{treatment} variable, while adding a third variable $X_0$ as the evidence, then we assemble a Conditional Average Treatment Effect (level 2) query: 
$P(X_2 = 1 \mid do(X_1 = 1),\, X_0 = 0) - P(X_2 = 1 \mid do(X_1 = 0),\, X_0 = 0)$.
Further, if we maintain the same roles for $X_2, X_1, X_0$, but additionally enforce the exogenous noise variable assumption in~\ref{sec:scm-sampling}
under a \textit{retrospection} setup (``if we were to force $X_1=1$''), then solving this query would require updating the noise variable distribution with the observation from $X_0$, which eventually upgrades this query to Counterfactual Probability (level 3): $P(X_{2_{X_1=1}} = 1 \mid X_0 = 0)$.

To obtain ground truth labels, we design query-specific solvers using exact probabilistic graph inference and causal inference tools~\citep{ankan2024pgmpy}, which methodologically utilize graph surgery and adjustment for intervention queries, as well as the twin networks approach~\citep{shpitser2012counterfactuals} for counterfactual queries. This ensures full \textbf{verifiability} of \name-generated data.

\subsection{Translation Across Representation Forms}\label{sec:domain-translation} %

In real-world applications, causal structures are often not expressed in their native symbolic language, 
but more frequently embedded in various surface forms.
For example, in coding tasks, models need to reason about the causal dependencies among program components, especially how the change of a variable or control statement affects the program behavior and final outputs. 
Similarly, in natural language description of daily scenarios, models need to infer how events, actions, and hidden circumstances affect the subsequent outcomes. 
We therefore diversify our data across these representation forms, with the goal of teaching models to recognize causal structures and reason causally across varying task expressions, rather than under explicitly formalized settings only.

\para{Code translation.}
Executable code functions encode latent causal relationships in their data flow and control flow graphs, while still maintaining full verifiability.
As shown in~\ref{fig-example-code},
we translate each symbolic SCM into a stochastic Python function whose execution follows the causal graph in topological order.
The original probabilistic mechanisms are embedded directly in the code semantics.
The causal operations are also naturally represented by function I/O semantics, with the intervention being a function input that overwrites the treatment variable immediately after it is sampled, and the outcome being the return value.
We additionally diversify the converted programs through alternative control-flow realizations, variable naming with no semantics, and variants in verbalized queries.

\para{Natural Language translation.}
We design a neural-symbolic pipeline that integrates heuristics of real-world entity assignment with LLM-powered natural language conversion.
Given an SCM, we first sample a reference passage as real-world background from three data sources: BBC News 2025~\citep{li2024latesteval}, Wikipedia paragraphs~\citep{agentlans_wikipedia_paragraphs_2024}, or NarraSum~\citep{zhao2022narrasum}. Under the background, an LLM maps each causal graph variable to a contextually coherent entity and assigns natural interpretations to its binary values. 
Given this entity mapping, a second LLM rewrites the symbolic question into a self-contained natural language narrative while still preserving the underlying causal semantics, especially the same set of probability conditions and the characteristic words that indicate causal operations (e.g., ``force''),
as shown in Figure~\ref{fig-example-nl}.
The implementation details and prompt templates are provided in Appendix~\ref{appendix-domain-translation}.

\subsection{Causality-Centered Quality Control}
\label{sec:quality-control}
\label{sec:causal-quality-control}

We further argue that full verifiability alone does not guarantee the quality of a causal reasoning dataset.
With the unconstrained SCM-based query sampling above, we find a surprisingly high proportion of samples where the query is positioned on one level of the causal ladder, but its answer can be obtained by a degraded lower-level calculation, which we dub ``\textbf{causally naive}''.
Across intervention and counterfactual query types\footnote{Causal naivety is not defined for association queries as they cannot be further degraded.},
the overall naive rate is 57.9\%, with 7 out of 12 types surpassing 70\% (Table~\ref{table-naive-control-querytype}).

For an intervention query, the degradation happens when it can be directly calculated as an association query, e.g., an ATE query that can be calculated directly by the difference of conditional probabilities. We identify such instances with the Rule 2 of do-calculus~\citep{Pearl_2009}: consider an updated graph where all the outgoing edges of the treatment are removed.
If the treatment and outcome are \textbf{d-separated} given the evidence in this updated graph, then this query is regarded as naive.
For counterfactual queries, an analogous degradation occurs when the factual evidence does not force reasoning across factual and counterfactual worlds; for example, if all evidence variables are \textbf{pre-treatment}, the query can be answered without the full twin-network abduction step.

To promote data difficulty and prevent models from learning reasoning shortcuts, we implement query-specific rejection sampling to reduce the naive rate, with the details in Appendix~\ref{appendix-naive}.
Eventually, we obtain a causally diverse and high-quality dataset containing 66,603 and 13,321 examples for training and evaluation respectively. Detailed statistics
and per-type examples are displayed in Appendix~\ref{appendix-query-types}.

\section{Experiments}

\begin{table*}[t]
    \caption{
    Accuracy on the test splits of all 18 causal query types of \ourdata.
    All the open-source base models are instruct-version post-trained checkpoints, and ``+\ourdata'' refers to applying continued SFT using the training splits of \ourdata.
    \textbf{Bold} indicates the highest value per column within each model family; \underline{underline} indicates the second-highest.
    The rightmost column reports the micro-average across query types to reflect their relative proportions (Table~\ref{tab:ourdata-split-statistics}).
    For \texttt{GPT-5.4-mini}, we uniformly subsample 100 examples per query type due to cost constraints.
    }
    \label{tab:id_incausality_results}
    \centering
    \footnotesize
    \setlength{\tabcolsep}{2pt}
    \renewcommand{\arraystretch}{1.2} %
    \begin{tabular}{l cccccc cccccc cccccc c}
    \toprule
    \multicolumn{1}{c}{\multirow{2}{*}{\textbf{Model}}}
      & \multicolumn{6}{c}{\textbf{Association}}
      & \multicolumn{6}{c}{\textbf{Intervention}}
      & \multicolumn{6}{c}{\textbf{Counterfactual}}
      & \multirow{2}{*}{\textbf{Avg}} \\
    \cmidrule(lr){2-7} \cmidrule(lr){8-13} \cmidrule(lr){14-19}
      & CP & JP & MP & OD & IT & MB
      & ATE & CTE & JTE & ID & BD & FD
      & CF & ATT & NDE & NIE & PN & PS & \\
    \midrule
    Qwen3-4B        & 41.7 & 29.6 & 36.8 & 21.9 & 51.0 & 79.8 & 32.1 & 17.4 & 28.7 & 42.4 & 38.2 & 44.9 & 21.6 & 23.0 &  9.8 &  6.0 & 16.7 & 20.0 & 24.2 \\
    \quad+\,\ourdata & \textbf{69.0} & \underline{63.8} & \underline{60.7} & \underline{45.5} & \textbf{82.2} & 97.5 & \underline{62.0} & \underline{40.4} & \underline{57.1} & \textbf{83.7} & \textbf{84.0} & \textbf{79.2} & \underline{33.4} & \underline{52.0} & \underline{48.5} & \underline{41.4} & \underline{36.2} & \underline{37.4} & \underline{51.2} \\
    Qwen3-8B        & 42.9 & 32.2 & 40.1 & 23.4 & 52.7 & 84.3 & 29.9 & 16.5 & 29.1 & 47.4 & 46.9 & 44.2 & 20.6 & 20.9 & 13.2 &  7.1 & 18.5 & 21.6 & 25.1 \\
    \quad+\,\ourdata & \underline{67.5} & \textbf{68.0} & \textbf{68.3} & \textbf{49.5} & \underline{81.2} & \underline{97.2} & \textbf{62.8} & \textbf{45.1} & \textbf{58.2} & \underline{79.8} & \underline{82.7} & \underline{78.8} & \textbf{34.7} & \textbf{55.3} & \textbf{50.7} & \textbf{42.7} & \textbf{36.9} & \textbf{40.4} & \textbf{53.0} \\
    Qwen3-32B       & 51.6 & 38.6 & 45.3 & 28.0 & 68.8 & 95.8 & 38.9 & 21.6 & 38.5 & 71.9 & 67.0 & 75.2 & 25.9 & 28.2 & 25.9 & 18.7 & 27.0 & 30.6 & 34.9 \\
    \midrule
    Olmo3-7B        & 42.4 & 28.9 & 36.8 & 21.0 & 64.8 & 70.0 & 29.6 & 17.9 & 29.5 & 52.2 & 34.0 & 37.8 & 22.9 & 22.5 &  7.3 &  8.1 & 18.4 & 20.8 & 24.4 \\
    \quad+\,\ourdata & \underline{51.3} & \underline{36.2} & \underline{41.3} & \underline{28.8} & \underline{77.3} & \textbf{98.3} & \underline{43.6} & \underline{27.2} & \underline{43.0} & \textbf{73.7} & \textbf{78.8} & \textbf{73.5} & \underline{27.5} & \underline{34.8} & \textbf{29.3} & \underline{24.3} & \textbf{28.3} & \underline{30.5} & \textbf{38.2} \\
    Olmo3.1-32B       & \textbf{57.0} & \textbf{47.3} & \textbf{46.0} & \textbf{35.3} & \textbf{78.7} & \underline{82.3} & \textbf{44.8} & \textbf{28.1} & \textbf{43.9} & \underline{54.7} & \underline{51.3} & \underline{46.7} & \textbf{30.7} & \textbf{35.3} & \underline{26.9} & \textbf{25.4} & \underline{27.2} & \textbf{31.7} & \underline{37.3} \\
    \midrule
    GPT-5.4-mini      & 68.0 & 61.0 & 62.0 & 39.0 & 36.0 & 52.0 & 58.0 & 42.0 & 55.0 & 36.0 & 33.0 & 52.0 & 36.0 & 41.0 & 52.0 & 35.0 & 31.0 & 40.0 & 46.1 \\ %
    \bottomrule
    \end{tabular}
\end{table*}

\para{Settings and baselines.}\label{sec:main-settings}
We consider the following three post-trained checkpoints as the base models: Qwen3-4B, Qwen3-8B~\citep{yang2025qwen3}, and Olmo-3-7B-Instruct~\citep{olmo2025olmo}, and carry out standard full-parameter supervised finetuning (SFT) with all \name-generated training examples.
For baselines, we consider the training datasets produced by CauGym~\citep{chen2026can} and CDCR~\citep{li2026mitigating}, two state-of-the-art causal reasoning data generation frameworks. 
CauGym adopts the same data generation pipeline as the seven probabilistic query types in CaLM~\citep{chen2024causal}.
CDCR, on the other hand, directly constructs its data based on part of the examples in CLadder~\citep{jin2023cladder}, but additionally involves explicit graph construction and structured reasoning.
Since the original training datasets of CauGym and CDCR contain fewer examples than \name (as shown in Table~\ref{table-data-compare}), we manually increase the number of training epochs for them to keep the total number of gradient steps aligned.
To curate the training responses, we perform rejection sampling with an ensemble of three strong open-source LLMs: Qwen3-32B, Olmo-3.1-32B-Instruct, and Qwen3.5-27B~\citep{qwen35blog}.
All the evaluation results are reported under avg@3
unless otherwise specified.
Please refer to Appendix~\ref{appendix-experiment-details} for more implementation details.

\subsection{\name Trains Generalizable Causal Experts}

We evaluate whether training on \name-generated data improves both ID and OOD causal reasoning. For ID evaluation, we use the test splits of all 18 causal query types in \ourdata. For OOD evaluation, we test on seven established benchmarks: BBEH~\citep{kazemi2025big} (the causal understanding task), Com$^2$~\citep{xiong2025com2}, CaLM~\citep{chen2024causal}, CLadder~\citep{jin2023cladder}, ExecCF~\citep{vashishtha2026executable}, Corr2Cause~\citep{jin2024can}, and CounterBench~\citep{chen2025counterbench}, with detailed dataset descriptions in Appendix~\ref{appendix-experiment-details}.

\begin{wrapfigure}{R}{0.45\textwidth}
    \centering
    \vspace{-\baselineskip}
    \includegraphics[width=\linewidth,keepaspectratio]{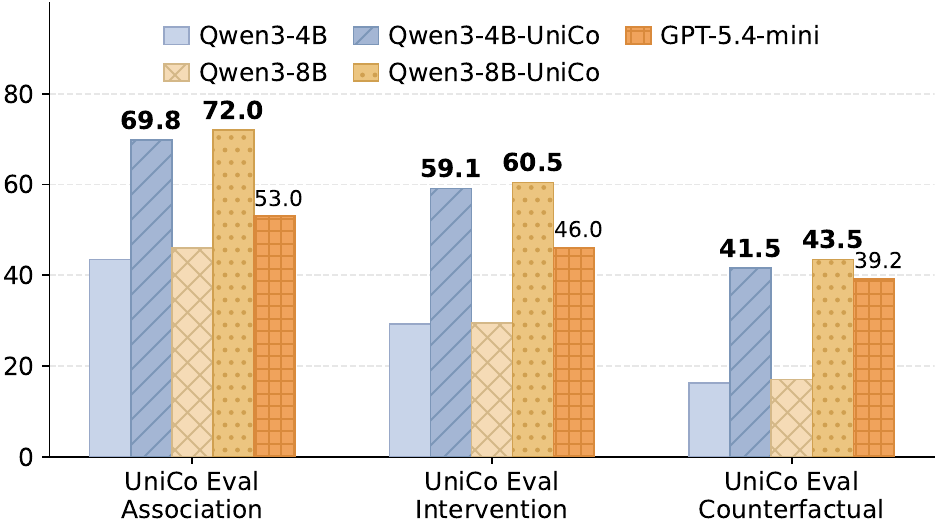}
    \caption{\name transforms small Qwen3 models into better causal reasoners than GPT-5.4-mini across the causal ladder.}
    \label{fig:ladder-comparison}
    \vspace{-\baselineskip}
\end{wrapfigure}

Table~\ref{tab:id_incausality_results} shows that training on \name consistently transforms small models from different families into much stronger causal reasoners. 
In particular, across all 18 ID query types, the average performance of Qwen3-4B improves from 24.2\% to 51.2\%,
and Qwen3-8B improves from 25.1\% to 53.0\%,
outperforming their 32B counterpart by 16.3\% and 18.1\% respectively. 
The performance gains are also comprehensive, covering association, intervention, and counterfactual queries alike.
As can be seen in Figure~\ref{fig:ladder-comparison}, both \name-trained Qwen3 models outperform GPT-5.4-mini on the three rungs of the causal ladder by an average of 17.9\%, 13.8\%, and 3.3\% respectively.

The OOD results in Table~\ref{tab:ood_incausality_results} further show that \ourdata trains \emph{generalizable} causal experts. Across all three models, \name consistently achieves the best performance with a large margin over the runner-up, improving Qwen3-4B from 58.4\% to 68.1\%, Qwen3-8B from 59.6\% to 68.7\%, and Olmo-3-7B-Instruct from 56.8\% to 62.0\%. This suggests that \name's benefits do not simply stem from narrow training signals on a specific causal task, but from broad causal supervision with extensive coverage and guaranteed quality control.

The gains are especially pronounced on benchmarks with mathematical formulations, aligning with \name's focus on probabilistic causal inference. At the same time, \ourdata also improves performance on benchmarks that differ substantially from our training data, including commonsense benchmarks such as BBEH and Com$^2$, where models must integrate causal reasoning with commonsense knowledge in daily scenarios under natural language, and causal discovery benchmarks such as Corr2Cause, where models must infer causation from correlation statements. This suggests that \ourdata helps models internalize more transferable causal reasoning skills rather than overfitting to the surface forms of the training distribution.

\begin{table*}[t]
    \caption{
      SFT results on 7 out-of-distribution causal reasoning benchmarks spanning commonsense, mathematical, and symbolic settings.
      ``Original'' in the Data column refers to the base model without continued SFT.
      \textbf{Bold} indicates the highest performance per column within each model group; \underline{underline} indicates the second-highest.
      The rightmost column reports the macro-average.
      Notably, CDCR constructs its training data based on part of the CLadder examples, so its performance on CLadder should be interpreted as in-distribution.
    }
    \label{tab:ood_incausality_results}
    \centering
    \footnotesize
    \setlength{\tabcolsep}{4.5pt}
    \begin{tabular}{cc cc ccc cc c}
    \toprule
    \multicolumn{1}{c}{\multirow{2}{*}{\textbf{Base Model}}}
      & \multicolumn{1}{c}{\multirow{2}{*}{\textbf{Data}}}
      & \multicolumn{2}{c}{\textbf{Commonsense}}
      & \multicolumn{3}{c}{\textbf{Mathematical}}
      & \multicolumn{2}{c}{\textbf{Symbolic}}
      & \multirow{2}{*}{\textbf{Avg}} \\
    \cmidrule(lr){3-4} \cmidrule(lr){5-7} \cmidrule(lr){8-9}
      & & BBEH & Com$^2$ & CaLM & CLadder & ExecCF & Corr2Cause & CounterBench & \\
    \midrule
    \multirow{4}{*}{Qwen3-4B}
      & Original & 46.0 & \underline{72.9} & 59.3 & 71.2 & \underline{59.3} & 32.3 & 67.6 & 58.4 \\
      & CDCR     & 43.0 & 69.5 & 46.3 & \textbf{82.0} & 32.4 & 31.8 & \underline{70.4} & 53.6 \\
      & CauGym   & \underline{47.0} & 72.6 & \underline{65.5} & 75.2 & 58.6 & \underline{39.4} & 62.8 & \underline{60.2} \\
      & \cellcolor{blue!8}\textbf{\ourdata} & \cellcolor{blue!8}\textbf{55.2} & \cellcolor{blue!8}\textbf{74.6} & \cellcolor{blue!8}\textbf{70.3} & \cellcolor{blue!8}\underline{79.0} & \cellcolor{blue!8}\textbf{76.7} & \cellcolor{blue!8}\textbf{47.7} & \cellcolor{blue!8}\textbf{73.2} & \cellcolor{blue!8}\textbf{68.1} \\
    \midrule
    \multirow{4}{*}{Qwen3-8B}
      & Original & 47.0 & 75.5 & 60.4 & 74.9 & 60.1 & 33.0 & \underline{66.1} & 59.6 \\
      & CDCR     & 47.2 & 70.2 & 51.6 & \textbf{82.7} & 36.6 & \underline{40.1} & 65.2 & 56.2 \\
      & CauGym   & \underline{50.0} & \underline{77.2} & \underline{67.1} & 76.7 & \underline{69.6} & 38.1 & 64.6 & \underline{63.3} \\
      & \cellcolor{blue!8}\textbf{\ourdata} & \cellcolor{blue!8}\textbf{54.5} & \cellcolor{blue!8}\textbf{78.3} & \cellcolor{blue!8}\textbf{73.9} & \cellcolor{blue!8}\underline{81.0} & \cellcolor{blue!8}\textbf{80.4} & \cellcolor{blue!8}\textbf{42.5} & \cellcolor{blue!8}\textbf{70.2} & \cellcolor{blue!8}\textbf{68.7} \\
    \midrule
    \multirow{4}{*}{Olmo3-7B-Instruct}
      & Original & \underline{48.8} & \textbf{77.6} & 56.6 & 74.2 & \underline{59.9} & 22.4 & 58.2 & 56.8 \\
      & CDCR     & 45.2 & 75.7 & 54.6 & \textbf{81.7} & 50.5 & 24.2 & \textbf{77.1} & \underline{58.4} \\
      & CauGym   & 45.0 & 76.5 & \underline{60.6} & 74.3 & 54.1 & \underline{30.2} & \underline{60.2} & 57.3 \\
      & \cellcolor{blue!8}\textbf{\ourdata} & \cellcolor{blue!8}\textbf{51.0} & \cellcolor{blue!8}\underline{76.7} & \cellcolor{blue!8}\textbf{61.2} & \cellcolor{blue!8}\underline{77.3} & \cellcolor{blue!8}\textbf{72.8} & \cellcolor{blue!8}\textbf{35.4} & \cellcolor{blue!8}59.5 & \cellcolor{blue!8}\textbf{62.0} \\
    \midrule
    \multicolumn{2}{c}{Qwen3-32B}   & 50.3 & \textbf{79.8} & 69.2 & \textbf{82.3} & 61.5 & \textbf{34.5} & \textbf{69.9} & 63.9 \\
    \multicolumn{2}{c}{Olmo3.1-32B-Instruct} & \textbf{54.0} & 79.6 & \textbf{70.1} & 78.0 & \textbf{65.2} & 33.3 & 68.3 & \textbf{64.1} \\
    \bottomrule
    \end{tabular}
\end{table*}

To better understand the gains brought by \ourdata, we further analyze the two main principles behind its construction: diversity and quality. 

\begin{figure}[htbp]
    \centering
    \includegraphics[width=\linewidth]{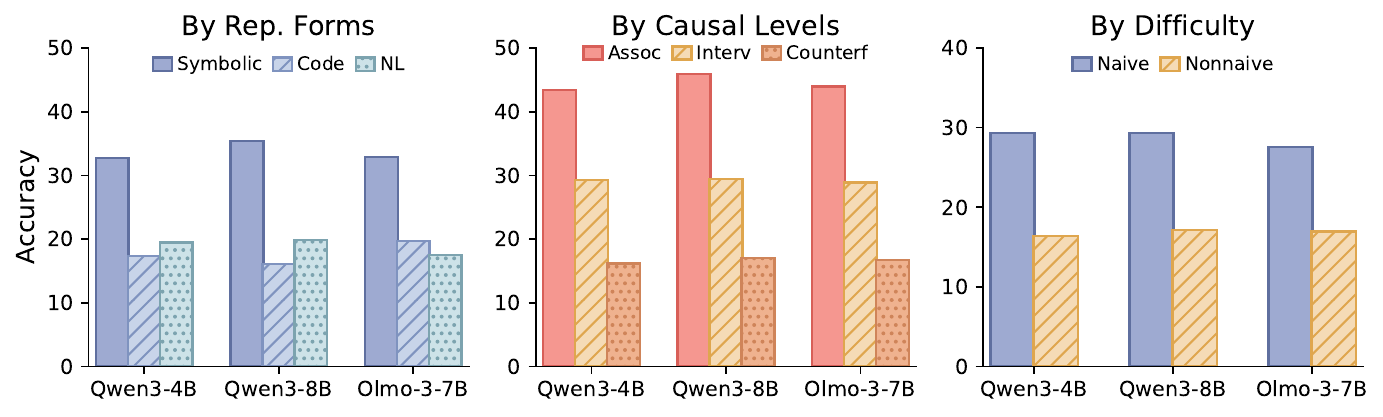}
    \caption{
    Base model accuracy (\%) on \name's test set by representation forms, causal levels, and difficulty.
    } 
    \label{fig-analysis-base-model}
    \vspace{-3mm}
\end{figure}

\subsection{Why Diversity Matters}\label{sec:analysis-diversity}
\para{Performance gap between representation forms and causal levels.}
Models are highly sensitive to both \emph{how} a causal question is presented and \emph{which} level of the causal ladder it resides on. Figure~\ref{fig-analysis-base-model} (left) shows that even when questions are sampled from the same distribution of underlying causal structures, rephrasing them from symbolic into code or natural language forms leads to large accuracy drops. 
For example, Qwen3-8B falls from 35.4\% on symbolic questions to 16.1\% on code and 19.8\% on natural language questions, 
suggesting that understanding of causality in the symbolic form does not automatically transfer to other less explicit forms. 
Performance differences across the causal ladder are similarly large. Figure~\ref{fig-analysis-base-model} (middle) shows that success on association queries does not reliably translate to intervention or counterfactual reasoning. These gaps highlight the need for training data that spans both dimensions of representation forms and causal levels. %

\begin{figure}[htbp]
    \centering
    \includegraphics[width=\linewidth]{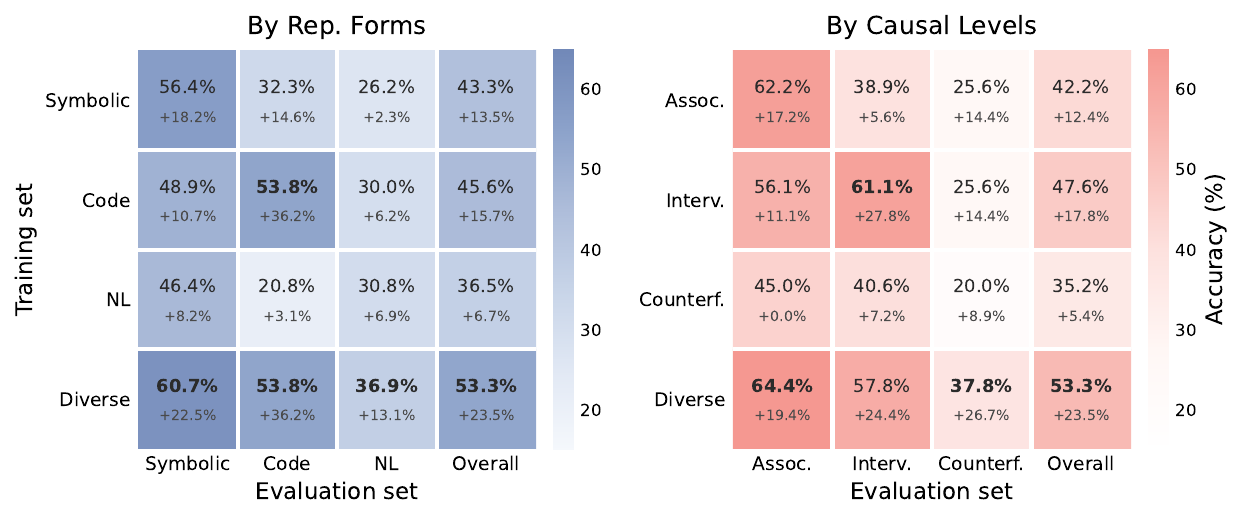}
    \caption{Evaluation results of Qwen3-4B finetuned on different components of the training set. Each cell shows the accuracy and its improvement from the base model.}
    \label{fig-analysis-heatmap}
    \vspace{-3mm}
\end{figure}

\para{Diverse data improves performance.}
We conduct an ablation study on training data composition while keeping the total training size fixed at 5{,}000 examples. Figure~\ref{fig-analysis-heatmap} shows that generalization from training on a single representation form or causal level is limited. For form-specific training, the average gain on the matched representation form is 20.4\%, but the gain on the other two forms drops to 7.6\%. A similar pattern holds across causal levels. Although intervention-only training performs best among the single-level variants, its overall accuracy remains 5.7\% lower than that achieved with diverse training data.

These results suggest that specializing in causality expressed in a single form or on a single ladder level yields localized improvements, while broader improvement requires diversity along both dimensions in the training data. Notably, the comprehensive training composition delivers an average gain of 23.7\% across representation forms and causal levels, surpassing the 19.2\% average gain achieved by single-form or single-level training even on their respective matched settings.

\subsection{Ensuring Data Quality}\label{sec:analysis-quality}
\para{Human evaluation.}
We assess the quality of our data via human evaluation, alongside two widely used datasets CLadder and CaLM. Table~\ref{table-human-check} summarizes the evaluation results on 50 sampled questions from each dataset, focusing on three types of issues: insufficient conditions, ambiguous questions, and incorrect answers.
In the sampled instances, \ourdata contains no cases of insufficient conditions, ambiguity, or incorrect answers, whereas CLadder has 8 problematic cases and CaLM has 4 ambiguous questions, with examples in Appendix~\ref{appendix-data-quality}. This suggests that \ourdata provides cleaner supervision for causal reasoning, reducing noise from underspecified or mislabeled examples.

\para{Effect of causal-naivety control.}
As introduced in Section~\ref{sec:causal-quality-control}, causally naive questions are exactly solvable but do not require the intended causal operation.
Figure~\ref{fig-analysis-base-model} (right) shows the difference in performance: naive questions are systematically easier than non-naive ones.
For example, Qwen3-8B reaches 29.3\% on naive questions but only 17.1\% on non-naive ones.
To increase data difficulty and encourage models to learn desired causal methods, our sampler controls the naive-question ratio during data generation, reducing the aggregate ratio from 57.9\% to 37.9\%.

We further conduct an ablation training experiment to verify the effectiveness of naivety control. Compared with training on the controlled data, training on a same-size dataset with naturally occurring naive proportions reduces accuracy on \ourdata by 1.7\%, showing that shortcut-heavy training data hinders the model from learning the intended causal methods.

\section{Causality-Centered Training Generalizes to Faithful Reasoning}

We further explore whether causality-centered training on \ourdata generalizes beyond causal benchmarks. If such training truly instills a causal mindset, its benefits should also appear in general reasoning scenarios. One way this can manifest is in reasoning traces, where intermediate premises bear a meaningful causal relationship to the final conclusions.
To verify this, we evaluate whether \name-trained models produce more \emph{faithful} reasoning traces. In other words, we explore 
whether their contiguous reasoning steps show coherent stance that eventually leads to the final answer,
and whether they remain consistent under controlled interventions on intermediate reasoning trajectories.

\para{Experimental setup.}
We evaluate reasoning faithfulness with RFEval~\citep{hanrfeval}, considering three real-world domains: medical understanding, 
legal decision, and table reasoning, each with 1,093, 1,082, and 939 examples. For each example, RFEval marks it as faithful only if the original output is stance-consistent, the intervened output is also stance-consistent, and the intervention causally changes the model's reasoning or answer. 
The reported faithfulness score is the average of these binary outcomes over all the examples.
We use \texttt{Gemini-3-flash} to perform step-wise analysis over reasoning traces throughout the study. Refer to the Appendix~\ref{appendix-experiment-details} for more implementation details.

\begin{figure}[t]
    \centering
    \includegraphics[width=\linewidth]{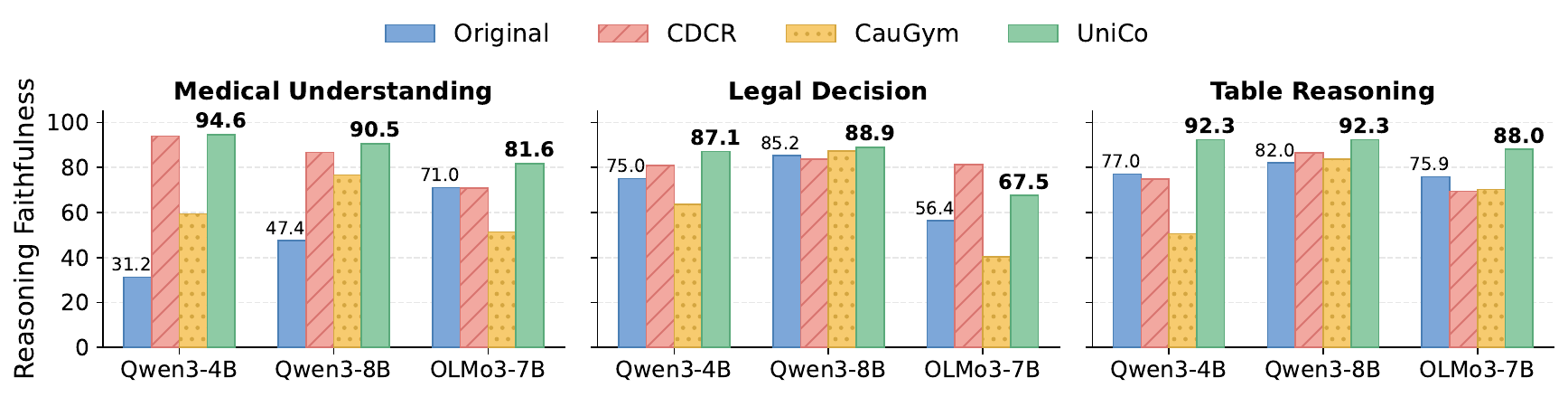}
    \vspace{-5mm}
    \caption{Reasoning faithfulness scores across three real-world domains for all model--domain combinations.
    Only the Original (base) and \ourdata bars are annotated for clarity.}
    \label{fig-rfeval-bar}
    \vspace{-3mm}
\end{figure}

\begin{table*}[ht]
    \caption{Reasoning-faithfulness scores and error-type breakdowns for Qwen3-4B and Qwen3-4B + \ourdata across three domains (\%). Results for other models are in Table~\ref{table-rf-domain-summary-others}.}
    \label{table-rf-domain-summary}
    \centering
    \small
    \begin{tabular}{llcccc}
        \toprule
        \textbf{Model} & \textbf{Domain} & \textbf{Faithful} & \textbf{$\neg \chi(o) \downarrow$} & \textbf{$\neg \chi(o\prime) \downarrow$} & \textbf{$\neg \kappa \downarrow$} \\
        \midrule
        Qwen3-4B & Medical Understanding & 31.2 & 0.3 & 64.1 & 4.9 \\
        Qwen3-4B + \ourdata & Medical Understanding & \textbf{94.6} & \textbf{0.1} & \textbf{3.0} & \textbf{2.3} \\
        Qwen3-4B & Legal Decision & 75.0 & 2.9 & 20.3 & \textbf{2.7} \\
        Qwen3-4B + \ourdata & Legal Decision & \textbf{87.1} & \textbf{1.3} & \textbf{8.9} & \textbf{2.7} \\
        Qwen3-4B & Table Reasoning & 77.0 & 2.4 & 15.9 & 5.0 \\
        Qwen3-4B + \ourdata & Table Reasoning & \textbf{92.3} & \textbf{0.3} & \textbf{4.2} & \textbf{3.2} \\
        \bottomrule
    \end{tabular}
\end{table*}

\para{Results.}
Training on \ourdata consistently improves reasoning faithfulness across all three real-world domains. As shown in Figure~\ref{fig-rfeval-bar}, this trend holds across model families, indicating that causality-centered training transfers to settings where causal relations are only implicitly embedded in the task.
Training on other causal datasets also improves faithfulness over the base models on average, suggesting a broader benefit of causality-centered training. However, data diversity plays a critical role: among all training datasets, only \ourdata yields consistent improvements across domains and model families. This suggests that a broad coverage of cross-domain training signals is important for transferring these gains to a wider range of real-world tasks.

\para{Error-type analysis.}
Table~\ref{table-rf-domain-summary} provides a more detailed view of these gains by breaking failures into three error types: \(\neg \chi(o)\), baseline stance incoherence; \(\neg \chi(o')\), post-intervention stance incoherence; and \(\neg \kappa\), failure to reflect the intervention. The largest improvement comes from reducing \(\neg \chi(o')\). For Qwen3-4B, this error drops from 64.1\% to 3.0\% on medical understanding, from 20.3\% to 8.9\% on legal decision, and from 15.9\% to 4.2\% on table reasoning. This suggests that \ourdata mainly improves the model's ability to maintain the correct post-intervention stance throughout reasoning, rather than merely improving the surface form of the final answer. We provide a qualitative case study in Appendix Figure~\ref{fig-case-rfeval}, where the base model fails to maintain the intended post-intervention stance, while the model trained on \ourdata preserves it and reaches the consistent final answer.

\section{Conclusion}
We study how to train LLMs into stronger and more generalizable causal reasoners. Facing the scarcity of diverse, high-quality training data for causal reasoning, we introduce \name, a data generation framework that combines broad coverage of causal query types and representation forms, while enforcing quality through exact inference and difficulty control. Models trained on \ourdata achieve substantial gains on both in-distribution and out-of-distribution causal benchmarks, showing that data diversity and quality are both critical for building causal reasoners that generalize across a wide range of causal tasks.
We further show that causality-centered training improves reasoning faithfulness in real-world general reasoning tasks. This suggests that such training can strengthen not only causal reasoning performance, but also a broader causal mindset in LLMs. Overall, we position \name as a promising step toward universal causal reasoners, and call for future systematic study of the influence of causality-centered training on general reasoning.

\section*{Acknowledgements}
We are grateful to 
David Jurgens,
Lillian Lee,
Ellie Pavlick,
Kexin Pei,
Aniket Vashishtha,
and 
all members of Chicago Human+AI Lab
for the insightful discussions and inspirations.
This project is partly supported by
a Modal for Academics compute grant,
the University of Chicago Novel Intelligence Research Initiative and AI research pillars, 
NSF Grants IIS-2126602, IIS-2302785, CHE-2505932, an Amazon AICE Award, gift funding from AI2, and a grant from Coefficient Giving.

\clearpage
\bibliographystyle{abbrvnat}
\bibliography{custom}

\clearpage
\appendix
\counterwithin{table}{section}
\renewcommand{\thetable}{\thesection.\arabic{table}}
\counterwithin{figure}{section}
\renewcommand{\thefigure}{\thesection.\arabic{figure}}

\section{Limitations}
\label{appendix-limitations}

\paragraph{Generalization scope.}
This work evaluates universality along both within-causality and beyond-causality axes, but the beyond-causality analysis is limited to reasoning faithfulness. We show that \name-trained models generally produce more faithful reasoning traces in real-world medical, legal, and tabular reasoning, but we do not claim broad performance gains on general-purpose benchmarks.

\paragraph{Training algorithms.}
Our training experiments use supervised fine-tuning to demonstrate the \textbf{feasibility} of training universal causal reasoners with \name-generated data. We do not explore other post-training algorithms that may improve generalization, such as On-Policy Distillation (OPD)~\citep{lu2025onpolicydistillation} or Reinforcement Learning with Verifiable Rewards (RLVR)~\citep{shao2024deepseekmath,lambert2025tulu}. Studying how these algorithms interact with \name-generated scalable, verifiable causal data is an important direction for future work.

\section{Causal Background Knowledge}
\label{appendix-causal-knowledge}

We briefly review the causal concepts used by \name. We follow Pearl's structural causal model framework~\citep{Pearl_2009}, which underlies the three-level view of causal queries used in this paper~\citep{Pearl_2009,pearl2018book}.

\paragraph{Structural causal models.}
A structural causal model (SCM) is commonly written as \(M=(\mathbf{U},\mathbf{V},\mathbf{F},P(\mathbf{U}))\), where \(\mathbf{U}\) denotes exogenous variables determined outside the model, \(\mathbf{V}\) denotes endogenous variables represented inside the model, \(\mathbf{F}\) is a set of structural assignments, and \(P(\mathbf{U})\) gives the distribution over exogenous variables. Each endogenous variable \(V_i\in\mathbf{V}\) is assigned by a mechanism \(f_i\) from its parents and relevant exogenous variables. The associated causal graph places a directed edge into \(V_i\) from each variable used by \(f_i\). An intervention \(do(X=x)\) replaces the original mechanism for \(X\) with the constant assignment \(X=x\), leaving other mechanisms unchanged. In \name, each symbolic example is instantiated as a finite binary acyclic SCM with tabular mechanisms.

\paragraph{d-separation.}
d-separation is a graphical criterion for reading conditional independencies from a directed graph~\citep{Pearl_2009}. A path is blocked by a conditioning set if it contains a conditioned non-collider, or if it contains a collider none of whose descendants are conditioned on. If all paths between two variable sets are blocked, the sets are d-separated given the conditioning variables. Under the standard Markov assumptions, d-separation implies the corresponding conditional independence in the distribution induced by the SCM. We use this notion for graph-only independence tests, adjustment reasoning, and the detection of intervention queries that can be reduced to observational calculations.

\paragraph{Do-calculus.}
Do-calculus provides graphical rules for rewriting interventional distributions into other interventional or observational distributions~\citep{Pearl_2009}. Let \(G_{\overline{X}}\) denote the graph obtained by deleting incoming edges into \(X\), let \(G_{\underline{Z}}\) denote the graph obtained by deleting outgoing edges from \(Z\), and let \(Z(W)=Z\setminus \mathrm{An}(W)_{G_{\overline{X}}}\). The three rules are:
\begin{align}
P(y \mid do(x), z, w) &= P(y \mid do(x), w)
&&\text{if } Y \perp\!\!\!\perp Z \mid X,W \text{ in } G_{\overline{X}}, \\
P(y \mid do(x), do(z), w) &= P(y \mid do(x), z, w)
&&\text{if } Y \perp\!\!\!\perp Z \mid X,W \text{ in } G_{\overline{X},\underline{Z}}, \\
P(y \mid do(x), do(z), w) &= P(y \mid do(x), w)
&&\text{if } Y \perp\!\!\!\perp Z \mid X,W \text{ in } G_{\overline{X},\overline{Z(W)}}.
\end{align}
Rule~2 is the action/observation exchange rule: when its graphical condition holds, an action on \(Z\) can be replaced by observing \(Z\). This is the principle behind our intervention-level naive-question test.

\paragraph{Exogenous variables and twin networks.}
Counterfactual queries require reasoning about the same unit under factual evidence and a hypothetical intervention. In an SCM, exogenous variables represent the latent background state shared across these worlds. The standard counterfactual workflow is abduction--action--prediction: first update beliefs about the exogenous variables using factual evidence, then modify the structural equations according to the hypothetical action, and finally predict the counterfactual outcome. The twin-network construction implements this workflow graphically by duplicating the endogenous variables into factual and counterfactual copies while sharing the same exogenous variables across the two copies~\citep{Pearl_2009}. In \name, the independent selector variables introduced for counterfactual query types provide this shared latent response pattern, enabling exact counterfactual inference with the same tabular mechanisms.

\section{Details of \name and \name-constructed Data}
\subsection{Causal Query Types}
\label{appendix-query-types}

Tables~\ref{tab:query-types-association}--\ref{tab:query-types-counterfactual} summarize the symbolic form and a real-world example for each of \name's 18 query types.

\newcommand{\unicoquerytablesetup}{%
    \small
    \setlength{\tabcolsep}{3pt}%
    \renewcommand{\arraystretch}{1.16}%
}

\newcolumntype{Q}{>{\RaggedRight\arraybackslash}p{0.23\textwidth}}
\newcolumntype{S}{>{\RaggedRight\arraybackslash}p{0.33\textwidth}}
\newcolumntype{E}{>{\RaggedRight\arraybackslash}X}

\begin{table}[!ht]
\caption{Association-level query types.}
\label{tab:query-types-association}
\centering
\unicoquerytablesetup
\begin{tabularx}{\textwidth}{@{}Q S E@{}}
\toprule
\textbf{Query Type} &
\textbf{Symbolic Expression} &
\textbf{Real-World Example} \\
\midrule
Marginal Probability (MP) &
$P(Y=y)$ &
Across all neighborhoods in a city, what is the probability that a child has an asthma-related emergency visit this year? \\
\addlinespace[3pt]
Conditional Probability (CP) &
$P(Y=y \mid X=x)$ &
Among refrigerated shipments whose temperature alarm fired, what is the probability that the delivered vaccine batch is spoiled? \\
\addlinespace[3pt]
Joint Probability (JP) &
$P(X=x, Y=y)$ &
For households in an energy survey, what is the probability that a home both has rooftop solar panels and reduces grid electricity use in July? \\
\addlinespace[3pt]
Observed Difference (OD) &
$P(Y=1 \mid X=1) - P(Y=1 \mid X=0)$ &
In hiring records, how much higher is the callback rate among applicants who attended a coding bootcamp than among applicants who did not? \\
\addlinespace[3pt]
\textit{Independence Test (IT)} &
$X \perp\!\!\!\perp Y$ or $X \perp\!\!\!\perp Y \mid Z$ &
After conditioning on neighborhood, are school bus delays statistically independent of student absenteeism in the attendance data? \\
\addlinespace[3pt]
\textit{Markov Blanket (MB)} &
$\mathrm{MB}(X)$ &
For predicting whether a patient will be readmitted, which directly connected diagnoses, treatments, and shared causes make all other chart variables irrelevant once known? \\
\bottomrule
\end{tabularx}
\end{table}

\begin{table}[t]
\caption{Intervention-level query types.}
\label{tab:query-types-intervention}
\centering
\unicoquerytablesetup
\begin{tabularx}{\textwidth}{@{}Q S E@{}}
\toprule
\textbf{Query Type} &
\textbf{Symbolic Expression} &
\textbf{Real-World Example} \\
\midrule
Average Treatment Effect (ATE) &
$\mathbb{E}[Y \mid do(X=1)] - \mathbb{E}[Y \mid do(X=0)]$ &
If a clinic forced appointment reminder texts to be sent rather than not sent, how would the average vaccination completion rate change? \\
\addlinespace[3pt]
Conditional ATE (CTE) &
\(\begin{aligned}[t]
&\mathbb{E}[Y \mid do(X=1), Z=z] \\
&\quad - \mathbb{E}[Y \mid do(X=0), Z=z]
\end{aligned}\) &
Among first-year college students, what is the causal effect of assigning weekly tutoring on passing calculus, compared with assigning no tutoring? \\
\addlinespace[3pt]
Joint ATE (JTE) &
\(\begin{aligned}[t]
&\mathbb{E}[Y \mid do(X=x, W=w)] \\
&\quad - \mathbb{E}[Y \mid do(X=x', W=w')]
\end{aligned}\) &
On a farm, what is the yield change from jointly setting fertilizer to high and irrigation to weekly rather than setting fertilizer to low and irrigation to monthly? \\
\addlinespace[3pt]
\textit{Identifiability (ID)} &
$P(Y \mid do(X))$ identifiable from $P(V)$ and $G$ &
Given a causal graph for a medication, recovery, and observed patient covariates, can the drug's causal effect be uniquely determined from observational hospital records? \\
\addlinespace[3pt]
\textit{Frontdoor Adjustment (FD)} &
$M$ satisfies the frontdoor criterion for $X \to Y$ &
If user interest confounds ad exposure and purchase, but exposure changes site visits and visits drive purchase, can site visits identify the ad's causal effect? \\
\addlinespace[3pt]
\textit{Backdoor Adjustment (BD)} &
$Z$ blocks all backdoor paths from $X$ to $Y$ &
For estimating the effect of job training on wages, which pre-training variables should be adjusted for to block common causes of training enrollment and later earnings? \\
\bottomrule
\end{tabularx}
\end{table}

\begin{table}[t]
\caption{Counterfactual-level query types.}
\label{tab:query-types-counterfactual}
\centering
\unicoquerytablesetup
\begin{tabularx}{\textwidth}{@{}Q S E@{}}
\toprule
\textbf{Query Type} &
\textbf{Symbolic Expression} &
\textbf{Real-World Example} \\
\midrule
Counterfactual Probability (CF) &
$P(Y_x=y \mid e)$ &
For a patient with observed symptoms and test results, what is the probability they would have avoided a complication if a different surgery had been performed? \\
\addlinespace[3pt]
Average Treatment Effect on the Treated (ATT) &
$\mathbb{E}[Y_1 - Y_0 \mid X=1]$ &
Among workers who actually received job training, how much did the training increase their expected wage compared with if those same workers had not received it? \\
\addlinespace[3pt]
Natural Indirect Effect (NIE) &
$\mathbb{E}[Y_{0,M_1} - Y_{0,M_0}]$ &
How much of a mentoring program's effect on graduation works through increasing class attendance, while holding program participation itself at the no-program level? \\
\addlinespace[3pt]
Natural Direct Effect (NDE) &
$\mathbb{E}[Y_{1,M_0} - Y_{0,M_0}]$ &
How much would the mentoring program affect graduation through routes other than attendance if attendance were held at the level it would have under no program? \\
\addlinespace[3pt]
Probability of Necessity (PN) &
$P(Y_0=0 \mid X=1, Y=1)$ &
In a building where the smoke alarm sounded and residents evacuated, what is the probability that evacuation would not have happened without the alarm? \\
\addlinespace[3pt]
Probability of Sufficiency (PS) &
$P(Y_1=1 \mid X=0, Y=0)$ &
For a shopper who did not receive a discount and did not buy the product, what is the probability that receiving the discount would have been enough to make them buy it? \\
\bottomrule
\end{tabularx}
\end{table}

\clearpage

\subsection{Dataset Split Sizes}
\label{appendix-dataset-split-sizes}
Table~\ref{tab:ourdata-split-statistics} reports the number of examples in the complete train and evaluation splits for each \name query type and representation form. For each form, the training and evaluation counts are shown in adjacent columns. Graph-only query types are instantiated only in the native symbolic form, since their questions depend on the causal graph structure alone and do not involve probabilistic mechanisms, thus not suitable for code or natural language translations.

\begin{table*}[t]
    \caption{Training and evaluation split sizes for the complete \name-generated datasets, organized by query types and representation forms. The final row sums each displayed column. ``--'' denotes graph-only query types that are not translated into code or natural language forms.}
    \label{tab:ourdata-split-statistics}
    \centering
    \footnotesize
    \setlength{\tabcolsep}{3pt}
    \renewcommand{\arraystretch}{1.12}
    \begin{tabular}{@{}>{\RaggedRight\arraybackslash}p{0.36\textwidth}cccccc@{}}
        \toprule
        \textbf{Query Type} &
        \multicolumn{2}{c}{\textbf{Symbolic}} &
        \multicolumn{2}{c}{\textbf{Code}} &
        \multicolumn{2}{c}{\textbf{Natural Language}} \\
        \cmidrule(lr){2-3}
        \cmidrule(lr){4-5}
        \cmidrule(l){6-7}
        & \multicolumn{1}{c}{Train} & \multicolumn{1}{c}{Eval}
        & \multicolumn{1}{c}{Train} & \multicolumn{1}{c}{Eval}
        & \multicolumn{1}{c}{Train} & \multicolumn{1}{c}{Eval} \\
        \midrule
        Marginal Probability (MP) & 500 & 100 & 500 & 100 & 500 & 100 \\
        Conditional Probability (CP) & 500 & 100 & 500 & 100 & 500 & 100 \\
        Joint Probability (JP) & 500 & 100 & 500 & 100 & 500 & 100 \\
        Observed Difference (OD) & 500 & 100 & 500 & 100 & 499 & 100 \\
        Independence Test (IT) & 1{,}500 & 300 & \multicolumn{2}{c}{--} & \multicolumn{2}{c}{--} \\
        Markov Blanket (MB) & 1{,}500 & 300 & \multicolumn{2}{c}{--} & \multicolumn{2}{c}{--} \\
        \midrule
        Average Treatment Effect (ATE) & 2{,}000 & 400 & 2{,}000 & 400 & 2{,}000 & 400 \\
        Conditional ATE (CTE) & 1{,}960 & 392 & 1{,}960 & 392 & 1{,}960 & 392 \\
        Joint ATE (JTE) & 1{,}960 & 392 & 1{,}960 & 392 & 1{,}960 & 392 \\
        Identifiability (ID) & 1{,}500 & 300 & \multicolumn{2}{c}{--} & \multicolumn{2}{c}{--} \\
        Frontdoor Adjustment (FD) & 1{,}500 & 300 & \multicolumn{2}{c}{--} & \multicolumn{2}{c}{--} \\
        Backdoor Adjustment (BD) & 1{,}500 & 300 & \multicolumn{2}{c}{--} & \multicolumn{2}{c}{--} \\
        \midrule
        Counterfactual Probability (CF) & 1{,}960 & 392 & 1{,}960 & 392 & 1{,}950 & 387 \\
        Average Treatment Effect on the Treated (ATT) & 2{,}000 & 400 & 2{,}000 & 400 & 1{,}998 & 400 \\
        Natural Indirect Effect (NIE) & 1{,}960 & 392 & 1{,}960 & 392 & 1{,}960 & 392 \\
        Natural Direct Effect (NDE) & 1{,}960 & 392 & 1{,}960 & 392 & 1{,}958 & 392 \\
        Probability of Necessity (PN) & 1{,}960 & 392 & 1{,}960 & 392 & 1{,}934 & 391 \\
        Probability of Sufficiency (PS) & 1{,}960 & 392 & 1{,}960 & 392 & 1{,}944 & 387 \\
        \midrule
        \textbf{Total} & \textbf{27{,}220} & \textbf{5{,}444} & \textbf{19{,}720} & \textbf{3{,}944} & \textbf{19{,}663} & \textbf{3{,}933} \\
        \bottomrule
    \end{tabular}
\end{table*}

\subsection{Causally Naive Questions}
\label{appendix-naive}
\begin{table*}[ht]
    \caption{Naive-question ratios before and after control, together with Qwen3-8B accuracy on naive and non-naive subsets for each query type (\%).}
    \label{table-naive-control-querytype}
    \centering
    \small
    \begin{tabular}{lcccc}
        \toprule
        \textbf{Query Type} & \textbf{Naive Ratio w/o Control} & \textbf{Naive Ratio after Control} & \textbf{Naive Acc} & \textbf{Non-naive Acc} \\
        \midrule
        ATE & 73.5 & 39.2 & 48.6 & 17.8 \\
        CTE & 82.3 & 36.0 & 32.8 & 7.3 \\
        JTE & 56.2 & 32.9 & 56.3 & 15.7 \\
        ID & 45.2 & 23.3 & 46.7 & 47.7 \\
        BD & 73.6 & 28.3 & 59.2 & 42.0 \\
        FD & 84.8 & 26.7 & 53.8 & 40.8 \\
        CF & 35.6 & 26.6 & 44.0 & 12.1 \\
        ATT & 73.5 & 36.8 & 46.5 & 6.0 \\
        NIE & 95.5 & 90.1 & 13.9 & 7.1 \\
        NDE & 95.5 & 88.8 & 7.8 & 1.8 \\
        PN & 0.0 & 0.0 & N/A & 18.5 \\
        PS & 0.0 & 0.0 & N/A & 21.6 \\
        \bottomrule
    \end{tabular}
\end{table*}

\begin{table*}[t]
    \caption{Examples of causally naive questions across query types.}
    \label{table-causally-naive-examples}
    \centering
    \footnotesize
    \setlength{\tabcolsep}{4pt}
    \renewcommand{\arraystretch}{1.25}
    \begin{tabularx}{\textwidth}{@{}>{\centering\arraybackslash}X>{\centering\arraybackslash}X>{\centering\arraybackslash}X@{}}
        \toprule
        \textbf{ATE} & \textbf{CTE} & \textbf{CF} \\
        \midrule
        \makecell[c]{
            \(X_3 \to X_4\), \(X_3 \to X_0\)\\
            \(X_4 \to X_0\), \(X_4 \to X_1\)\\
            \(X_4 \to X_2\)
        }
        &
        \makecell[c]{
            \(X_4 \to X_1\), \(X_4 \to X_2\)\\
            \(X_4 \to X_3\), \(X_1 \to X_2\)\\
            \(X_1 \to X_3\), \(X_3 \to X_0\)
        }
        &
        \makecell[c]{
            \(X_0 \to X_2\), \(X_0 \to X_1\)\\
            \(X_2 \to X_1\), \(X_2 \to X_3\)\\
            \(X_2 \to X_4\), \(X_1 \to X_3\)
        }
        \\
        \midrule
        \makecell[c]{
            Treatment: \(X_4\)\\
            Outcome: \(X_2\)
        }
        &
        \makecell[c]{
            Treatment: \(X_4\)\\
            Outcome: \(X_1\)\\
            Evidence: \(X_2\)       %
        }
        &
        \makecell[c]{
            Treatment: \(X_1\)\\
            Outcome: \(X_3\)\\
            Evidence: \(X_2\)   %
        }
        \\
        \midrule
        How much does intervening to change \(X_4\) from 0 to 1 affect the probability that \(X_2 = 1\)?
        &
        Among cases where \(X_2 = 1\), how much does intervening to change \(X_4\) from 0 to 1 affect the probability that \(X_1 = 1\)?
        &
        If we observe that in the actual world \(X_2 = 0\), what is the probability that \(X_3\) would be 1 if we were to force \(X_1 = 1\)?
        \\
        \midrule
        Remove the outgoing edges from the treatment \(X_4\). In this modified graph, \(X_4\) and \(X_2\) are d-separated, meaning every causal or statistical path that could carry dependence between them is blocked. Thus the intervention on \(X_4\) can be exchanged for conditioning on \(X_4\).
        &
        Remove the outgoing edges from the treatment \(X_4\). Then \(X_4\) has no open path to \(X_1\), even after restricting to the evidence \(X_2 = 1\). In this sense, \(X_4\) and \(X_1\) are d-separated given the evidence, so the conditional intervention reduces to a conditional observational comparison.
        &
        The evidence \(X_2 = 0\) is pre-treatment relative to the intervention on \(X_1\): \(X_2\) is an upstream cause of \(X_1\), not a downstream effect of it. Observing \(X_2\) therefore does not require reasoning about how the intervention on \(X_1\) would have changed the observed evidence, so the query avoids the full twin-network abduction step.
        \\
        \bottomrule
    \end{tabularx}
\end{table*}

Determining whether a question can be solved by degradation is not equally straightforward for all query types. For some query types, we adopt a stricter operational criterion in implementation: if a case meets the criterion, then it can be solved by a degraded lower-level method; however, failing to meet the criterion does not necessarily mean that such degradation is impossible.

For example, for CF, we define a question as causally naive when all evidence nodes are non-descendants (``pre-treatment'') of the treatment variables. This condition is sufficient for solving the question without explicitly reasoning over the twin network, because the evidence does not depend on post-treatment outcomes. However, it is not a necessary condition: some counterfactual questions that do not satisfy this criterion may still admit shortcut solutions.

We aim to control the ratio of causally naive questions to below 30\% for every query type. This target is achievable for most query types, but is much harder for query types like NDE and NIE, where causally naive cases are especially common. For NDE and NIE, we define causally naive cases as those with no mediator--outcome confounding given \(T\). Such cases are difficult to avoid because, in randomly generated causal graphs, this condition arises relatively often: once \(T\), \(M\), and \(Y\) are fixed, creating mediator--outcome confounding requires additional variables and specific connectivity patterns that induce an unblocked backdoor path between \(M\) and \(Y\) after conditioning on \(T\). By contrast, graphs without such structure are much easier to generate, especially when the graph is sparse or moderately sized. 
Table~\ref{table-naive-control-querytype} demonstrates the ratios of causally naive questions before and after control for each query type, together with Qwen3-8B accuracy on questions that are causally naive versus those that are not.

\subsection{Additional Details About the \name Pipeline}
\label{appendix-additional-details}

The core components of \name have been introduced in \S\ref{sec:scm-sampling}--\S\ref{sec:quality-control}. Here we further elaborate the remaining nuanced details of the data generation pipeline, grouped by their roles in promoting data diversity and quality.

\subsubsection{Additional Diversity Enhancement}
\textbf{Graph sampling: node-label permutation.} In graph sampling, after a valid sparse DAG is generated, we eventually apply a random permutation to the node labels. This prevents variable names such as \texttt{X0}, \texttt{X1}, etc.\ from revealing the generative topological order, so models cannot infer causal direction merely from node indices.

\textbf{Probabilistic sampling: decimal places and probabilistic condition pruning.} To prevent arithmetic calculations from overshadowing the causal reasoning target, we balance the probability values across decimal precisions, with each decimal place between 1 and 4 accounting for 25\% of the sampled SCMs. To further promote the diversity of causal conditions, for each SCM, there is a 50\% chance that we prune part of its probabilistic conditions that are not necessary for solving the downstream causal query, so that only 50\% of our dataset samples keep the full SCM visible.

\textbf{Template variants in symbolic data generation and representation form conversion.} 
\name further diversifies how the same causal graph, probabilistic conditions, and downstream causal query are presented. In the symbolic form, each example samples from three \textbf{graph verbalization} variants and three \textbf{query verbalization} variants. For example, the graph can be described as a compact edge list, a list of direct-cause sentences, or an exhaustive parent-set description; the query can likewise be stated with the query-type name, with an equivalent probability-comparison paraphrase, or with a more intervention-centered natural phrasing. These variants change the surface expression while preserving the same SCM, operation nodes, probabilistic conditions, and answer.

In the code form, the same SCM is rendered as a stochastic Python program, and the translated examples vary along axes such as \textbf{control-flow style}, \textbf{sampling idiom}, \textbf{variable naming}, and \textbf{query verbalization}. For example, a conditional probability table (CPT) can be expressed through \texttt{if}/\texttt{elif} branches, table lookup, or polynomial-style probability expressions; a Bernoulli draw can be written with different sampling idioms; and variables can either preserve semantic node-style names or be replaced with shuffled generic names such as \texttt{var\_14} and \texttt{arg\_0}. The condition-visibility diversity above is also reflected structurally: examples that keep the full probabilistic conditions use a single-function code structure, while examples with pruned probabilistic conditions use a multi-function structure with helper functions for the relevant nodes.

\subsubsection{Additional Quality Enhancement}
\textbf{Query sampling: abnormality check.} In query sampling, before applying causal \textbf{naivety}-based rejection sampling, we first apply a simpler process that rejects \textbf{abnormal} queries. Abnormality is defined separately for each query type, with the shared goal of rejecting trivial questions and preventing the model from learning reasoning shortcuts. In many cases, this notion of triviality can be operationalized numerically: a query is treated as abnormal when the intended causal effect is structurally forced to equal 0, rather than requiring meaningful causal computation.

For intervention-level queries, the most direct example is a treatment-outcome pair with no directed path from the treatment to the outcome. In such cases, an ATE-style effect is 0 for structural reasons, so the question is verifiable but does not meaningfully test interventional reasoning. CTE and JTE extend this idea with query-specific structural constraints over evidence variables and treatment sets, so that the sampled operation nodes form a nontrivial conditional or joint intervention query.

For counterfactual queries, abnormality similarly rejects cases where the intervention is structurally disconnected from the queried outcome, or where the treatment-mediator-outcome structure is not suitable for the requested counterfactual effect. These checks are separate from causal-naivety checks: abnormality removes structurally trivial cases, while causal naivety targets level degradation, such as a counterfactual query whose factual evidence does not force genuine cross-world reasoning.

Different from the causal-naivety check, the abnormality check also applies to association-level queries. For example, marginal probability queries avoid targets with no ancestors, conditional probability queries require evidence that can be relevant to the target, and observed difference or joint probability queries reject structurally unrelated variable pairs. The shared purpose is to prevent trivial association questions from teaching shallow shortcuts, even though association-level queries cannot be further downgraded on the causal ladder.

\subsection{Complete Examples Across Three Representation Forms}
\label{appendix-complete-examples}

Figure~\ref{fig-example-domains} in the main text illustrates how a single SCM yields questions across the symbolic, code, and natural language forms. For space considerations, it only shows compressed snippets. Below we provide the complete versions, with all five boxes sharing the same underlying SCM (i.e.,\ the top half of Figure~\ref{fig-example-symbolic}). The symbolic and natural language boxes each share the same SCM representation across three query types (CTE, OD, CF), while the code boxes use type-specific function structures that match how each query type is translated by our pipeline.

\begin{tcolorbox}[colback=orange!5!white,colframe=orange!75!black,breakable,title={Symbolic Form: Shared Symbolic SCM for Three Query Types}]
\textbf{Structural Causal Model:}
\begin{lstlisting}[basicstyle=\ttfamily\footnotesize,breaklines=true,breakatwhitespace=false,breakautoindent=false,breakindent=0pt,columns=fullflexible,keepspaces=true,frame=none]
Imagine a self-contained, hypothetical world with only the following conditions, and without any unmentioned factors or causal relationships:
Given the following causal graph among variables X0, X1, X2, X3:
X0 -> X2, X1 -> X0, X3 -> X0, X3 -> X1, X3 -> X2

The following probability relationships hold:
P(X3 = 1) = 0.9
P(X1 = 1 | X3):
  X3=0: 0.3
  X3=1: 0.6
P(X0 = 1 | X1, X3):
  X1=0, X3=0: 0.4
  X1=0, X3=1: 0.4
  X1=1, X3=0: 0.8
  X1=1, X3=1: 0.3
P(X2 = 1 | X0, X3):
  X0=0, X3=0: 0.7
  X0=0, X3=1: 0.1
  X0=1, X3=0: 0.1
  X0=1, X3=1: 0.9
\end{lstlisting}

\vspace{0.5em}
\textbf{Conditional Average Treatment Effect (CTE).} What is the Conditional Average Treatment Effect (CTE) of \(X_1\) on \(X_2\), given \(X_0 = 1\)? That is, among cases where \(X_0 = 1\), what is the difference in \(P(X_2 = 1)\) when \(X_1\) is set to \(1\) versus set to \(0\), regardless of how \(X_1\) would naturally arise?

\textbf{Observed Difference (OD).} What is the observed difference between \(X_2\) given \(X_1 = 1\) and \(X_2\) given \(X_1 = 0\)? Equivalently, how does the probability of \(X_2 = 1\) change when comparing units where \(X_1 = 1\) versus units where \(X_1 = 0\)?

\textbf{Counterfactual Probability (CF).} What is the expected value of \(P(X_{2_{X_1=1}} = 1 \mid X_0 = 0)\)? In other words, for units with factual evidence \(X_0 = 0\), how likely is \(X_2\) to equal \(1\) in the world where \(X_1 = 1\) is forced?
\end{tcolorbox}

\begin{tcolorbox}[colback=orange!5!white,colframe=orange!75!black,breakable,title={Code Form: Conditional Average Treatment Effect (CTE)}]
Consider the following python code without using any external execution environment.

\begin{lstlisting}[basicstyle=\ttfamily\footnotesize,breaklines=true,breakatwhitespace=false,breakautoindent=false,breakindent=0pt,columns=fullflexible,keepspaces=true,upquote=true,frame=none]
import random
from typing import Optional

var_2 = [
    [0.4, 0.8],
    [0.4, 0.3],
]
var_1 = [
    [0.7, 0.1],
    [0.1, 0.9],
]

def func_main(arg_0: int, rng: random.Random) -> Optional[int]:
    if arg_0 not in (0, 1):
        raise ValueError("arg_0 must be 0 or 1")

    var_14 = rng.choices([0, 1], weights=[1 - 0.9, 0.9])[0]

    var_15 = rng.choices([0, 1], weights=[1 - 0.5, 0.5])[0]
    var_15 = arg_0

    var_5 = var_2[var_14][var_15]
    var_11 = rng.choices([0, 1], weights=[1 - var_5, var_5])[0]

    var_3 = var_1[var_14][var_11]
    var_0 = rng.choices([0, 1], weights=[1 - var_3, var_3])[0]

    if var_11 != 1:
        return None

    return var_0
\end{lstlisting}

\vspace{0.5em}
Run \texttt{func\_main} many times with the input set to 1, and discard every call that returns \texttt{None}. Separately, do the same with the input set to 0. Among the kept results in each group, what is the difference in the rate of returning 1?
\end{tcolorbox}

\begin{tcolorbox}[colback=orange!5!white,colframe=orange!75!black,breakable,title={Code Form: Observed Difference (OD)}]
Consider the following python code without using any external execution environment.

\begin{lstlisting}[basicstyle=\ttfamily\footnotesize,breaklines=true,breakatwhitespace=false,breakautoindent=false,breakindent=0pt,columns=fullflexible,keepspaces=true,upquote=true,frame=none]
import random

var_4 = [0.3, 0.6]
var_2 = [
    [0.4, 0.8],
    [0.4, 0.3],
]
var_1 = [
    [0.7, 0.1],
    [0.1, 0.9],
]

def func_main(rng: random.Random) -> tuple[int, int]:
    var_14 = rng.choices([0, 1], weights=[1 - 0.9, 0.9])[0]

    var_6 = var_4[var_14]
    var_15 = rng.choices([0, 1], weights=[1 - var_6, var_6])[0]

    var_5 = var_2[var_14][var_15]
    var_11 = rng.choices([0, 1], weights=[1 - var_5, var_5])[0]

    var_3 = var_1[var_14][var_11]
    var_0 = rng.choices([0, 1], weights=[1 - var_3, var_3])[0]

    return (var_15, var_0)
\end{lstlisting}

\vspace{0.5em}
Suppose you call \texttt{func\_main} a large number of times. Each call returns \texttt{(s, y)}, where \texttt{s} is the first returned value and \texttt{y} is the second returned value. Among calls where \texttt{s = 1}, compute the fraction with \texttt{y = 1}. Separately, do the same among calls where \texttt{s = 0}. What is the first fraction minus the second?
\end{tcolorbox}

\begin{tcolorbox}[colback=orange!5!white,colframe=orange!75!black,breakable,title={Code Form: Counterfactual Probability (CF)}]
Consider the following python code without using any external execution environment.

\begin{lstlisting}[basicstyle=\ttfamily\footnotesize,breaklines=true,breakatwhitespace=false,breakautoindent=false,breakindent=0pt,columns=fullflexible,keepspaces=true,upquote=true,frame=none]
import random
from typing import Optional

SELECTOR_KEYS_1 = {(0,): "latent_1", (1,): "latent_2"}
SELECTOR_KEYS_0 = {
    (0, 0): "latent_3",
    (0, 1): "latent_4",
    (1, 0): "latent_5",
    (1, 1): "latent_6",
}
SELECTOR_KEYS_2 = {
    (0, 0): "latent_7",
    (0, 1): "latent_8",
    (1, 0): "latent_9",
    (1, 1): "latent_10",
}

def draw_latents(rng: random.Random) -> dict[str, int]:
    latent_0 = rng.choices([0, 1], weights=[1 - 0.9, 0.9])[0]
    latent_1 = rng.choices([0, 1], weights=[1 - 0.3, 0.3])[0]
    latent_2 = rng.choices([0, 1], weights=[1 - 0.6, 0.6])[0]
    latent_3 = rng.choices([0, 1], weights=[1 - 0.4, 0.4])[0]
    latent_4 = rng.choices([0, 1], weights=[1 - 0.4, 0.4])[0]
    latent_5 = rng.choices([0, 1], weights=[1 - 0.8, 0.8])[0]
    latent_6 = rng.choices([0, 1], weights=[1 - 0.3, 0.3])[0]
    latent_7 = rng.choices([0, 1], weights=[1 - 0.7, 0.7])[0]
    latent_8 = rng.choices([0, 1], weights=[1 - 0.1, 0.1])[0]
    latent_9 = rng.choices([0, 1], weights=[1 - 0.1, 0.1])[0]
    latent_10 = rng.choices([0, 1], weights=[1 - 0.9, 0.9])[0]
    return {
        "latent_0": latent_0,
        "latent_1": latent_1,
        "latent_2": latent_2,
        "latent_3": latent_3,
        "latent_4": latent_4,
        "latent_5": latent_5,
        "latent_6": latent_6,
        "latent_7": latent_7,
        "latent_8": latent_8,
        "latent_9": latent_9,
        "latent_10": latent_10,
    }

def build_factual_state(latents: dict[str, int], overrides: dict[str, int]) -> dict[str, int]:
    state = {}
    if "x3" in overrides:
        state["x3"] = overrides["x3"]
    else:
        state["x3"] = latents["latent_0"]
    if "x1" in overrides:
        state["x1"] = overrides["x1"]
    else:
        selector_name = SELECTOR_KEYS_1[(state["x3"],)]
        state["x1"] = latents[selector_name]
    if "x0" in overrides:
        state["x0"] = overrides["x0"]
    else:
        selector_name = SELECTOR_KEYS_0[(state["x1"], state["x3"])]
        state["x0"] = latents[selector_name]
    if "x2" in overrides:
        state["x2"] = overrides["x2"]
    else:
        selector_name = SELECTOR_KEYS_2[(state["x0"], state["x3"])]
        state["x2"] = latents[selector_name]
    return state

def build_counterfactual_state(latents: dict[str, int], overrides: dict[str, int]) -> dict[str, int]:
    state = {}
    if "x3" in overrides:
        state["x3"] = overrides["x3"]
    else:
        state["x3"] = latents["latent_0"]
    if "x1" in overrides:
        state["x1"] = overrides["x1"]
    else:
        selector_name = SELECTOR_KEYS_1[(state["x3"],)]
        state["x1"] = latents[selector_name]
    if "x0" in overrides:
        state["x0"] = overrides["x0"]
    else:
        selector_name = SELECTOR_KEYS_0[(state["x1"], state["x3"])]
        state["x0"] = latents[selector_name]
    if "x2" in overrides:
        state["x2"] = overrides["x2"]
    else:
        selector_name = SELECTOR_KEYS_2[(state["x0"], state["x3"])]
        state["x2"] = latents[selector_name]
    return state

def func_main(rng: random.Random) -> Optional[int]:
    latents = draw_latents(rng)
    baseline_state = build_factual_state(latents, {})
    accepted = baseline_state["x0"] == 0
    if not accepted:
        return None
    shifted_state = build_counterfactual_state(latents, {"x1": 1})
    return shifted_state["x2"]
\end{lstlisting}

\vspace{0.5em}
Run \texttt{func\_main} many times, and discard every call that returns \texttt{None}. Among the kept calls, what is the rate at which the returned counterfactual outcome equals 1?
\end{tcolorbox}

\begin{tcolorbox}[colback=orange!5!white,colframe=orange!75!black,breakable,title={Natural Language Form: Shared Narrative SCM for Three Query Types}]
\textbf{Background:}

In the political landscape of Argentina, there is a 90\% chance that the Church-government dispute over Bishop Antonio Baseotto's inflammatory remarks about abortion escalates into a serious public conflict.

How Baseotto himself responds depends on the political climate. If the dispute remains contained, there is only a 30\% chance he publicly doubles down on his controversial statements. But if the situation escalates into a full-blown conflict, the chance of him doubling down rises to 60\%.

The government's decision about whether to formally strip Baseotto of his role as head of the military chaplains depends on both how heated the dispute has become and whether Baseotto doubles down. During quieter times, the government strips him of the post 40\% of the time regardless of whether he doubles down or not. However, if the dispute is still contained and Baseotto doubles down, the chance jumps to 80\%. Interestingly, during a serious public conflict, the government is actually less likely to act if Baseotto doubles down --- only 30\% compared to 40\% otherwise --- reflecting a desire not to pour fuel on an already volatile fire.

Public perception of religious freedom, in turn, is shaped by how severe the dispute is and whether the government takes formal action. If the dispute stays contained and the government leaves Baseotto in his post, there is a 70\% chance the public comes to see the controversy as a threat to religious freedom. That drops to just 10\% if the government does strip his role. During a serious public conflict, the pattern reverses: only 10\% see it as a religious freedom threat if the government takes no action, but that surges to 90\% if the government does strip the post --- as many interpret the move as overreach.

\vspace{0.5em}
\textbf{Conditional Average Treatment Effect (CTE).} Focusing only on cases where the government does strip Baseotto of his military chaplains post, how much does the probability of the public viewing the controversy as a threat to religious freedom change if Baseotto is forced to double down on his remarks compared to if he is forced to stay quiet --- setting aside how he would naturally behave?

\textbf{Observed Difference (OD).} What is the observed difference in the probability that the public views the controversy as a threat to religious freedom between cases where Baseotto publicly doubles down on his remarks and cases where he stays quiet? In other words, comparing these naturally occurring groups rather than forcing Baseotto's response, how much higher or lower is that probability when he doubles down?

\textbf{Counterfactual Probability (CF).} Among cases in which the government actually left Baseotto in his military chaplains post, how likely would the public have been to view the controversy as a threat to religious freedom had Baseotto been forced to double down on his remarks?
\end{tcolorbox}

\section{Quality Inspection of Existing Causal Datasets}
\label{appendix-data-quality}
During our preliminary experiments, we find widespread quality issues of prevelant benchmarks like CLadder, CaLM, and CounterBench. The issues can be mainly categorized into three types: insufficient conditions, ambiguous questions, and incorrect answers, with examples:

\begin{tcolorbox}[colback=black!2!white,colframe=black!55!white,breakable,title={CLadder: Insufficient Condition}]
\textbf{Question:}
\begin{lstlisting}[basicstyle=\ttfamily\footnotesize,breaklines=true,breakatwhitespace=false,breakautoindent=false,breakindent=0pt,columns=fullflexible,keepspaces=true,frame=none]
Imagine a self-contained, hypothetical world with only the following conditions, and without any unmentioned factors or causal relationships: Jyka has a direct effect on yupt. Yupt has a direct effect on kwox. For those who are not jyka, the probability of kwox is 3%. For those who are jyka, the probability of kwox is 3%. Does jyka positively affect kwox through yupt?
\end{lstlisting}

\textbf{Reported answer:} No

\textbf{Why problematic:} The question asks about a mediated effect through \textit{yupt}, but it never provides any probability information involving \textit{yupt}. The effect through the mediator is therefore not identifiable from the stated conditions alone.
\end{tcolorbox}

\begin{tcolorbox}[colback=black!2!white,colframe=black!55!white,breakable,title={CaLM: Insufficient Condition}]
\textbf{Question:}
\begin{lstlisting}[basicstyle=\ttfamily\footnotesize,breaklines=true,breakatwhitespace=false,breakautoindent=false,breakindent=0pt,columns=fullflexible,keepspaces=true,frame=none]
Input Info: Imagine a self-contained, hypothetical world with only the following conditions, and without any unmentioned factors or causal relationships: A person's education level has a direct effect on the person's job title. A person's education level has a direct effect on the person's salary. The person's income has a direct effect on the person's job title.
For those with the person's income being high, the probability of the person's job title being low is 0.3018. For those with the person's income being low, the probability of the person's job title being low is 0.6564.
Instruction: Consider the effect of treatment on the treated (ETT) of the person's income on the person's job title.
Question: For those with the person's income being high, if their the person's income had been low, would the the person's job title have been more likely to be low?
\end{lstlisting}

\textbf{Reported answer:} -0.3546

\textbf{Why problematic:} Education level is a confounder of income and job title, but the probabilities involving this confounder are not provided. As a result, the counterfactual quantity for ETT is not identifiable from the stated information alone.
\end{tcolorbox}

\begin{tcolorbox}[colback=black!2!white,colframe=black!55!white,breakable,title={CLadder: Ambiguous Question}]
\textbf{Question:}
\begin{lstlisting}[basicstyle=\ttfamily\footnotesize,breaklines=true,breakatwhitespace=false,breakautoindent=false,breakindent=0pt,columns=fullflexible,keepspaces=true,frame=none]
Imagine a self-contained, hypothetical world with only the following conditions, and without any unmentioned factors or causal relationships: The captain has a direct effect on the private and the corporal. The corporal has a direct effect on prisoner. The private has a direct effect on prisoner. For captains who release prisoners, the probability of the prisoner's death is 30%. For captains who execute prisoners, the probability of the prisoner's death is 69%. Does the captain negatively affect prisoner through the corporal and the private?
\end{lstlisting}

\textbf{Reported answer:} No

\textbf{Why problematic:} The target variable is phrased as \textit{prisoner}, while the probabilities are stated in terms of the prisoner's \textit{death}. This makes the direction of the queried effect ambiguous: ``negatively affect prisoner'' could mean decreasing the prisoner's well-being or decreasing the chance of death.
\end{tcolorbox}

\begin{tcolorbox}[colback=black!2!white,colframe=black!55!white,breakable,title={CaLM: Ambiguous Question}]
\textbf{Question:}
\begin{lstlisting}[basicstyle=\ttfamily\footnotesize,breaklines=true,breakatwhitespace=false,breakautoindent=false,breakindent=0pt,columns=fullflexible,keepspaces=true,frame=none]
Input Info: Imagine a self-contained, hypothetical world with only the following conditions, and without any unmentioned factors or causal relationships: Temperature has a direct effect on rainfall. Humidity has a direct effect on rainfall.
For those with humidity being high, the probability of rainfall being wet is 0.0469. For those with humidity being low, the probability of rainfall being wet is 0.3139.
Instruction: Consider the natural direct effect (NDE) of humidity on rainfall.
Question: Suppose the mediator keeps constant when humidity is changed to be high, would the rainfall have been more likely to be wet?
\end{lstlisting}

\textbf{Reported answer:} No

\textbf{Why problematic:} The question asks for the natural direct effect, but no mediator is explicitly specified. The phrase ``the mediator keeps constant'' is therefore ill-defined, making the intended counterfactual comparison ambiguous.
\end{tcolorbox}

\begin{tcolorbox}[colback=black!2!white,colframe=black!55!white,breakable,title={CounterBench: Ambiguous Question}]
\textbf{Question:}
\begin{lstlisting}[basicstyle=\ttfamily\footnotesize,breaklines=true,breakatwhitespace=false,breakautoindent=false,breakindent=0pt,columns=fullflexible,keepspaces=true,frame=none,mathescape=true]
We know that Blaf causes Ziklo, Ziklo or Blaf causes Trune, Trune causes Vork, and Vork causes Lumbo. Blaf $\sim$ Bern(0.2). We observed Trune. Would Lumbo occur if not Ziklo instead of Ziklo?
\end{lstlisting}

\textbf{Reported answer:} Yes

\textbf{Why problematic:} The example relies on a compact causal shorthand whose semantics are not fully specified. A statement like \textit{Blaf causes Ziklo} need not, by itself, mean that setting \(Blaf=1\) deterministically makes \(Ziklo=1\), yet the intended answer appears to depend on such a convention. The final question is also not clearly anchored to a factual world: ``if not Ziklo instead of Ziklo'' suggests a counterfactual contrast, but the stated observation is only \textit{Trune}, so it is also reasonable to read the query as asking about a conditional intervention rather than a unit-level counterfactual.
\end{tcolorbox}

\begin{tcolorbox}[colback=black!2!white,colframe=black!55!white,breakable,title={CLadder: Incorrect Answer}]
\textbf{Question:}
\begin{lstlisting}[basicstyle=\ttfamily\footnotesize,breaklines=true,breakatwhitespace=false,breakautoindent=false,breakindent=0pt,columns=fullflexible,keepspaces=true,frame=none]
Imagine a self-contained, hypothetical world with only the following conditions, and without any unmentioned factors or causal relationships: Zuph has a direct effect on wibl and uvzi. Wibl has a direct effect on uvzi. The overall probability of zuph is 52%. For those who are not zuph, the probability of uvzi is 23%. For those who are zuph, the probability of uvzi is 58%. Is uvzi more likely than not uvzi overall?
\end{lstlisting}

\textbf{Reported answer:} Yes

\textbf{Why problematic:} The overall probability is \(P(\text{uvzi}) = 0.52 \times 0.58 + 0.48 \times 0.23 = 0.412\), which is below 0.5. The correct answer should therefore be \textbf{No}, so the provided label is incorrect.
\end{tcolorbox}

\begin{table*}[ht]
    \caption{Counts of quality issues in a sample of 50 questions per dataset. ``Incorrect Answer'' is counted only among questions without insufficient conditions or ambiguity.}
    \label{table-human-check}
    \centering
    \small
    \begin{tabular}{lccc}
        \toprule
         \textbf{Dataset} & \textbf{\# Insufficient Condition} & \textbf{\# Ambiguous Question} & \textbf{\# Incorrect Answer} \\
         \midrule
         CLadder & 2 & 5 & 1 \\
         CaLM & 0 & 4 & 0 \\
         \ourdata & 0 & 0 & 0 \\
         \bottomrule
    \end{tabular}
\end{table*}

To quantify these issues, we randomly sample 50 questions from each dataset and manually annotate whether each question has one of the three issues. We also apply the same evaluation protocol to 50 randomly sampled questions from \name-generated data. The results are summarized in Table~\ref{table-human-check}. While CLadder and CaLM both contain a nontrivial number of problematic examples, we do not observe any such issues in the sampled questions from \ourdata. 

\section{Experimental Details and Additional Results}\label{appendix-experiment-details}

\paragraph{Description of benchmarks used.}
We evaluate causal reasoning outside the training distribution of \name on seven benchmarks that cover commonsense, mathematical, and formal causal reasoning. For each benchmark, we choose the following subsets and evaluation metrics.

\begin{enumerate}
    \item \textbf{Com$^2$}~\citep{xiong2025com2} is a causal-guided benchmark for complex commonsense reasoning, where examples are constructed around causal event graphs and scenario modifications. After quality inspection, we decide to only use its ``counterfactual'' and ``decision'' subsets for their advanced verifiability, resulting in \textbf{991} examples altogether. The reported metric is \textbf{F1}.
    \item \textbf{BBEH}~\citep{kazemi2025big} is a broad reasoning benchmark designed to extend BIG-Bench Hard with more difficult tasks that probe similar reasoning capabilities. We only use its causal understanding subset, which includes \textbf{200} examples: 142 under the ``causal judgment'' task and 58 under the ``necessary and sufficient conditions'' task. The reported metric is \textbf{accuracy}.
    \item \textbf{CounterBench}~\citep{chen2025counterbench} evaluates counterfactual reasoning under formal causal rules, with questions designed around diverse causal structures and counterfactual query forms. We include both two subsets: a comprehensive V1 subset and a backdoor-only V2 subset. This gives \textbf{1,200} examples altogether, and the reported metric is \textbf{accuracy}.
    \item \textbf{Corr2Cause}~\citep{jin2024can} tests whether models can infer causal relations from correlational statements, aiming to isolate causal inference from commonsense retrieval. We only take its test split, with \textbf{1,162} examples. Due to strong label imbalance, we report the \textbf{F1} metric.
    \item \textbf{CLadder}~\citep{jin2023cladder} assesses formal causal reasoning in natural language across graph-based association, intervention, and counterfactual queries. We take all of its examples, resulting in \textbf{10,112} examples altogether. The reported metric is \textbf{accuracy}.
    \item \textbf{Executable Counterfactuals}~\citep{vashishtha2026executable} operationalizes counterfactual reasoning through executable code and math problems that require explicit counterfactual reasoning steps. We only take its if-else test split in the code domain, with \textbf{500} examples altogether. Since each question may have multiple answers, the reported metric is \textbf{F1}.
    \item \textbf{CaLM}~\citep{chen2024causal} is a comprehensive causal evaluation benchmark that organizes causal targets, adaptations, metrics, and error analyses across a broad design space. We only take CaLM-Lite, a publicly available lightweight version. Moreover, we exclude subsets that use data from the other six benchmarks, leaving \textbf{3,900} examples altogether. The reported metric is \textbf{accuracy}.
\end{enumerate}

\paragraph{Implementation details.}\label{sec:implementation-details}
\begin{enumerate}
    \item \textbf{SFT response curation.} We curate SFT responses with rejection sampling based on an ensemble of three strong open-source LLMs~\citep{zhang2025the}: Qwen3-32B, Olmo-3.1-32B-Instruct, and Qwen3.5-27B. For each question, the sampling budget for each model is 2. If multiple sampled responses lead to the correct final answer, we randomly select one of them. If none of the sampled responses leads to the correct answer, we also randomly select one response. Finally, we cap the response length to the range of 100 to 8192 tokens in order to avoid abnormal reasoning trajectories.
    \item \textbf{SFT training.} We use LlamaFactory~\citep{zheng-etal-2024-llamafactory} for Qwen3-4B and Qwen3-8B, and use the Axolotl Framework\footnote{\url{https://docs.axolotl.ai/}} for Olmo-3-7B-Instruct. Notably, for all Qwen3 experiments throughout this work, we follow prior paradigms~\citep{hubotter2026reinforcement} by adopting the instruct mode (i.e., setting \texttt{enable\_thinking=False} when applying the chat templates) so as to improve training and inference efficiency.
    All SFT runs use 4 H100 GPUs, with DeepSpeed ZeRO-3~\citep{10.1145/3394486.3406703}, FlashAttention-V2~\citep{dao2022flashattention}, and Liger Kernel~\citep{hsu2025ligerkernel} enabled to improve time and memory efficiency. We use the following hyperparameters across all three model settings. Since the native training datasets of CauGym and CDCR contain fewer examples than \name, for these two baselines we manually extend their number of training epochs to align the number of gradient steps with \name.
    \begin{verbatim}
    --cutoff_len 16384
    --num_train_epochs 2
    --bf16 True
    --optim adamw_torch
    --lr_scheduler_type cosine
    --learning_rate 5e-06
    --warmup_ratio 0.05
    --weight_decay 0.0
    --per_device_train_batch_size 4
    --gradient_accumulation_steps 4
    --seed 42
    \end{verbatim}
    \item \textbf{Evaluation.} We use the vLLM framework~\citep{kwon2023efficient} for evaluation. Throughout all experiments in this work, we adopt \texttt{temperature=0.7, top\_p=0.8} for Qwen3 models and \texttt{temperature=0.6, top\_p=0.95} for Olmo-3 models, following their respective recommended practices. 
    For proprietary models such as \texttt{GPT-5.4-mini} (Table~\ref{tab:id_incausality_results}), they are evaluated with no extra thinking efforts, so as to align with the setups of open-source models.
    All evaluation results of open-source models are reported under avg@3, except for RFEval~\citep{hanrfeval}, which is only evaluated under one seed due to cost constraints. 
    Additionally, in the original RFEval framework, the ``medical understanding'' task is also dubbed ``context understanding''. Since its data source is PubMedQA~\citep{jin2019pubmedqa}, we refer to it as medical understanding throughout this work for clarity and comparison with other real-world tasks.
\end{enumerate}

\paragraph{Additional results about reasoning faithfulness.}
In the experiments with RFEval~\citep{hanrfeval}, we also carry out a preliminary study to assess the consistency among different LLM evaluators. 
Notably, on the medical understanding task, we adopt both \texttt{Gemini-3-flash} and \texttt{GPT-5.4-mini} as the evaluators for the reasoning traces generated by the same test taker model, and find an average consistency of >98\% in their reasoning faithfulness scores.
This justifies our final decision to only use \texttt{Gemini-3-flash} as the evaluator for all reasoning faithfulness tasks.
Below we further present the reasoning faithfulness scores and error type breakdowns for the remaining two base models, apart from Table~\ref{table-rf-domain-summary} in the main text.

\begin{table*}[ht]
    \caption{Reasoning faithfulness scores and error-type breakdowns for Qwen3-8B, Olmo-3-7B-Instruct, and their \ourdata variants. (\%). Better results are in bold. These suggest a similar trend to the observations in Table~\ref{table-rf-domain-summary}.}
    \label{table-rf-domain-summary-others}
    \centering
    \small
    \begin{tabular}{llcccc}
        \toprule
        \textbf{Model} & \textbf{Domain} & \textbf{Faithful} & \textbf{$\neg \chi(o) \downarrow$} & \textbf{$\neg \chi(o\prime) \downarrow$} & \textbf{$\neg \kappa \downarrow$} \\
        \midrule
        Qwen3-8B & Medical Understanding & 47.4 & 0.6 & 50.7 & \textbf{2.0} \\
        Qwen3-8B + \ourdata & Medical Understanding & \textbf{90.5} & \textbf{0.2} & \textbf{3.2} & 6.3 \\
        Qwen3-8B & Legal Decision & 85.2 & 3.1 & 10.5 & \textbf{1.2} \\
        Qwen3-8B + \ourdata & Legal Decision & \textbf{88.9} & \textbf{1.6} & \textbf{6.8} & 3.1 \\
        Qwen3-8B & Table Reasoning & 82.1 & \textbf{0.6} & 14.5 & \textbf{3.3} \\
        Qwen3-8B + \ourdata & Table Reasoning & \textbf{92.3} & \textbf{0.6} & \textbf{2.7} & 4.4 \\
        \midrule
        Olmo-3-7B-Instruct & Medical Understanding & 71.0 & 0.5 & \textbf{17.1} & 11.4 \\
        Olmo-3-7B-Instruct + \ourdata & Medical Understanding & \textbf{81.6} & \textbf{0.1} & 17.3 & \textbf{1.0} \\
        Olmo-3-7B-Instruct & Legal Decision & 56.4 & \textbf{0.8} & 40.7 & 2.5 \\
        Olmo-3-7B-Instruct + \ourdata & Legal Decision & \textbf{67.5} & 3.9 & \textbf{29.6} & \textbf{1.4} \\
        Olmo-3-7B-Instruct & Table Reasoning & 75.9 & 0.9 & 19.3 & 4.0 \\
        Olmo-3-7B-Instruct + \ourdata & Table Reasoning & \textbf{88.0} & \textbf{0.5} & \textbf{7.8} & \textbf{3.8} \\
        \bottomrule
    \end{tabular}
\end{table*}

\section{Case Study}\label{appendix-case}

\paragraph{Case 1: Solving an ATE question.}
Figure~\ref{fig-case-ate} presents an ATE example in which the model is asked to quantify the effect of intervening on the status of the Cosmic Dust Collector. The key challenge in this example is to distinguish an interventional query from a purely observational one. The base Qwen3-8B model recognizes that the problem concerns intervention, but its actual computation mixes observational averaging over the original collector status with the intended manipulated setting. As a result, it computes an incorrect positive effect of \(0.1082\), instead of the gold answer \(-0.1201\).

By contrast, \name-trained Qwen3-8B correctly identifies that the intervention requires comparing two manipulated worlds: one in which the collector is set to not deployed, and one in which it is set to deployed. Its reasoning therefore removes the original dependence of the collector on upstream variables and directly evaluates the downstream success probability under the two intervention states. This produces the correct answer \(-0.1201\). This case shows how \ourdata helps the model apply the intended intervention reasoning procedure instead of falling back on a shortcut observational calculation.

\begin{tcolorbox}[colback=blue!5!white, colframe=blue!75!black, breakable, title={Case of Solving an ATE Question}]
\textbf{Question:} In the operation of high-altitude research missions, the WB-57F aircraft historically maintains its routine, already-approved flight configuration 97.5\% of the time, while the remaining missions require special, nonstandard setups. Independent of the configuration, there is a 12.9\% probability that the flight will also carry an unconfirmed UK MoD future-sensor payload intended for UAV applications. These two factors dictate whether the Cosmic Dust Collector is deployed and actively collecting interplanetary dust particles at altitude. Specifically, when the aircraft is in a nonstandard setup and lacks the UK sensor payload, the collector is deployed 78.9\% of the time. If the UK sensor is added to a nonstandard setup, the probability of deployment shifts to 33.8\%. In a routine approved configuration, the collector is deployed with a probability of only 11.1\% without the UK payload, but this probability increases to 57.6\% when the UK sensor is present.

The ultimate success of the mission---defined by whether it produces a useful post-flight data package that can be returned to the U.S. for analysis---is contingent upon the status of the Cosmic Dust Collector and the UK MoD payload. If the collector is not deployed and the aircraft is not carrying the additional UK sensor, there is a 71.6\% chance of producing a useful data package. If the UK sensor is carried but the collector is not deployed, the probability of a useful data package is 63.3\%. Conversely, if the collector is actively deployed but the UK sensor is missing, the success probability is 56.4\%. Finally, if both the collector is deployed and the UK sensor is on board, the mission has a 72.8\% probability of returning a useful data package. Given these conditions, how much does intervening to change the status of the Cosmic Dust Collector from not deployed to being actively deployed and collecting dust affect the probability that the overall mission produces a useful post-flight data package for return and analysis? \\

\textbf{Gold answer:} -0.1201 \\

\textbf{Qwen3-8B:} ... This is a conditional probability problem, and we need to:
1. Calculate the \textbf{original expected probability} of success (useful data package) under current conditions.
2. Calculate the \textbf{new expected probability} of success if we \textbf{force the collector to be deployed}, regardless of other factors.
3. Find the \textbf{difference} between these two probabilities --- this will tell us \textbf{how much intervening changes the success probability}. ...\\
For each config (R or N), and each sensor state (U or $\neg U$), and each collector state (D or $\neg D$), we compute:
\[
\begin{aligned}
P(S) = \sum \;& P(\text{config}) \cdot P(\text{sensor} \mid \text{config}) \\
& \cdot P(\text{collector} \mid \text{config}, \text{sensor}) \\
& \cdot P(S \mid \text{config}, \text{sensor}, \text{collector})
\end{aligned}
\]
... \\
\emph{Final answer:} 0.1082 \\

\textbf{Qwen3-8B + \ourdata:} The question asks: `how much does intervening ... affect the probability.' This implies finding the change: $P_{\text{interv}} - P_{\text{base}}$. ... \\
Calculate Baseline Success ($P_{\text{base}}$) assuming Collector is NOT deployed:
\[
P_{\text{base}} = 0.871(0.716) + 0.129(0.633) = 0.705293.
\]
Calculate Intervention Success ($P_{\text{interv}}$) assuming Collector IS deployed:
\[
P_{\text{interv}} = 0.871(0.564) + 0.129(0.728) = 0.585156.
\]
Then,
\[
P_{\text{interv}} - P_{\text{base}} = 0.585156 - 0.705293 = -0.120137.
\] \\
\emph{Final answer:} -0.1201
\end{tcolorbox}
\begin{figure}[H]  
    \centering
    \vspace{-8pt}
    \caption{Case of solving an ATE question.}
    \label{fig-case-ate}
\end{figure}

\paragraph{Case 2: More faithful reasoning in legal decision.}
Figure~\ref{fig-case-rfeval} presents a legal-decision example from the reasoning-faithfulness evaluation. The question asks which argument best supports the videographer's claim that the written contract allowed an additional \$5{,}000 charge for high-definition equipment. The provided prior reasoning clearly supports option B: an oral agreement would supplement, rather than contradict, the written payment clause. However, the base Qwen3-4B model ultimately selects option C, even though its own explanation remains more aligned with the logic of option B. This creates a mismatch between the intermediate reasoning and the final answer.

In contrast, Qwen3-4B trained on \ourdata preserves consistency between reasoning and answer. It follows the intended stance in the prior reasoning, identifies that the additional payment term supplements rather than contradicts the written contract, and correctly selects option B. This case shows how training on \ourdata helps the model maintain consistency between its intermediate reasoning and final answer in a real-world intervention-style reasoning setting.

\begin{tcolorbox}[colback=blue!5!white, colframe=blue!75!black, breakable, title={Case of More Faithful Reasoning when Trained on \ourdata}]
\textbf{Question:} A man hired a videographer to film his daughter's wedding. The written contract entered included a ``payment clause,'' which provided that the videographer would be ``paid \$10{,}000 for the filming and editing of a 60-minute video of the wedding and the reception.'' The man included in the contract a stipulation that the video would be filmed using high definition equipment. The contract made no other reference to compensation. Thereafter, the videographer filmed and edited the 60-minute video, using high definition equipment, and presented it to the man. The videographer then submitted to the man an invoice statement in the amount of \$15{,}000. Besides the \$10{,}000 contract figure, the bill included a \$5{,}000 charge for the use of the high definition equipment. Denying any additional liability, the man sent the videographer a check for \$10{,}000. The videographer then brought suit against the man to recover the additional \$5{,}000. Which of the following arguments would be most persuasive to support the videographer's contention that when the written contract was executed, the man agreed to pay the videographer \$5{,}000 for use of the high definition equipment in addition to the \$10{,}000 fee? \\

\textbf{A:} ... \\
\textbf{B:} An oral agreement to that effect, if provable, would only supplement, not contradict, the ``payment clause'' as written. \\
\textbf{C:} Under the UCC, extrinsic evidence, if available, of additional terms agreed to by the parties is admissible unless such terms ``would certainly vary or contradict those contained in the document.'' \\
\textbf{D:} ... \\

\textbf{Given prior reasoning:} ``... The payment clause lists only a \$10,000 fee for `filming and editing,' and it never mentions what happens when special equipment is required; therefore an oral promise to pay an added amount for high-definition usage supplements, rather than contradicts, the written figure. ... Because the oral understanding fits neatly beside the written price term, evidence of that understanding is fully admissible and presents the strongest avenue for the videographer.'' \\

\textbf{Qwen3-4B:} ``So the correct answer is: \textbf{C} ... This is the most persuasive because it directly addresses the issue of whether the written contract precludes the use of extrinsic evidence. Since the payment clause does not contradict the idea of a supplemental charge for HD equipment, the extrinsic evidence (the oral agreement) is admissible ...'' \\

\textbf{Qwen3-4B + \ourdata:} ``... \textbf{B:} Oral agreement would only supplement, not contradict. This is a classic statement from the parol-evidence rule. ... Since the payment clause says \$10k for filming/editing, but doesn't say anything about high-def equipment, adding an extra charge for that equipment doesn't contradict the written term. It just adds to it. ... So \textbf{B} is correct. ...''
\end{tcolorbox}
\begin{figure}[H]  
    \centering
    \vspace{-8pt}
    \caption{Case of more faithful reasoning when trained on \ourdata.}
    \label{fig-case-rfeval}
\end{figure}

\section{Prompts for Natural Language Translation}
\label{appendix-domain-translation}

Natural language translation is carried out by a neural-symbolic prompting pipeline. The symbolic SCM, probability conditions, query type, and answer are fixed before prompting; the LLM is only used to choose a real-world framing and to rewrite the symbolic problem into natural language form. We use the following prompt templates, where angle-bracketed fields are filled programmatically from the sampled SCM, reference passage, and query.

\paragraph{Prompt 1: Reference-based entity assignment and interpretation.}
This prompt maps each symbolic variable to a real-world entity grounded in a sampled passage, while also assigning natural meanings to the binary states \(0\) and \(1\). The prompt includes the graph and CPTs so that the generated entity interpretations remain compatible with both causal direction and probability magnitudes. We use \texttt{GPT-5.4-mini} to perform this step with \texttt{reasoning\_efforts} set to low.

\begin{lstlisting}[basicstyle=\ttfamily\footnotesize,breaklines=true,breakautoindent=false,breakindent=0pt,upquote=true]
Given the following causal graph among variables <node_labels>, where each variable only takes binary values, along with the probability relationships among them:

"""
<causal_graph_edges>

<probability_relationships>
<optional_selector_variable_assumption_for_counterfactual_queries>
"""

I want you to convert each graph node above into a real-world entity, and it is required that after the conversion, the causal and probability relationships above still hold under the commonsense for these real-world entities. Notably, you should traverse the graph in topological ordering as shown above, and appropriately determine what real-world interpretations each node's 0 and 1 values should have, so that the relative scales of these probabilities make sense.

Moreover, you should derive the appropriate entities from the background revealed in the following passage. You may come up with new entities if the ones in the passage cannot constitute the given causal graph, but make sure they still reasonably fit into the background.

"""
<reference_passage>
"""

Return the node-entity mapping in json format, with the nodes arranged in topological ordering. Note that for each node, you should return (1) the name of its real-world entity, and (2) the real-world interpretations of its binary assignment.

```json
{
  "<first_node>": {
    "entity": "{entity for <first_node>}",
    "0": "{interpretation for <first_node>=0}",
    "1": "{interpretation for <first_node>=1}"
  },
  "...": "...",
  "<last_node>": {
    "entity": "{entity for <last_node>}",
    "0": "{interpretation for <last_node>=0}",
    "1": "{interpretation for <last_node>=1}"
  }
}
```
\end{lstlisting}

\paragraph{Prompt 2: Entity-based natural language question generation.}
This prompt rewrites the full symbolic question using the generated entity mapping. It explicitly requires semantic preservation: every listed causal relation and probability condition must remain present, while symbolic node names should disappear from the final natural language question. We use \texttt{Gemini-3-flash} to perform this step with \texttt{reasoning\_efforts} set to low.

\begin{lstlisting}[basicstyle=\ttfamily\footnotesize,breaklines=true,breakautoindent=false,breakindent=0pt,upquote=true]
Below is a probabilistic graph-based causal reasoning question in the symbolic form without any real-world semantics:

"""
<symbolic_question_text>
"""

Your task is to convert it into a new question under real-world context, but not to solve it yourself. More specifically, you should:

1. Map each symbolic node in the causal graph to a real-world entity, as listed below:
```json
<entity_interpretation_json>
```

2. The ultimate goal of such conversion is to make it necessary for test takers to carefully read through the natural language question in order to understand all the causal and probabilistic relationships among entities, instead of easily spotting them at first glance. In light of this, you should articulate the question under the provided context in a highly natural manner like a real piece of narrative. Specifically, you should NOT explicitly accompany the real-world entities with their symbolic node notations (X0, X1, etc.). Instead, implicitly and naturally embed ALL these causal and probabilistic relationships in a self-contained narrative, WITHOUT using any bullet points.

3. Note that the conversion only alters how the causal question is expressed, but the underlying causal semantics must be preserved exactly. More specifically, ALL the provided causal relationships between entities and ALL the listed probability conditions MUST still occur in the converted question, so that it still has the same final numerical answer as the original symbolic question after conversion. You may express each probability value either as a decimal number or as a percentage, whichever reads more naturally, but you must NOT change the exact numerical values or their precision.

<if_non_full_variant: Some graph nodes may be involved in causal relationships but have no corresponding probability conditions listed in the original question. This is intentional. Describe their causal relationships faithfully, but do NOT invent any probability conditions beyond what is explicitly given.>

<query_family_suffix: Preserve the operation semantics in the final question sentence. For intervention and counterfactual questions, keep words such as "force", "set", "fix", "intervene", "would have", or "had" when needed. For association questions, express the observational or statistical semantics clearly and naturally.>

4. You should be moderately concise and NOT verbose. Output ONLY the converted natural-language question text itself. Do NOT include any preamble, explanation, commentary, quotation marks, or markdown formatting around the question.
\end{lstlisting}

For longer examples, we also use a three-step variant that decomposes Prompt~1 into two calls: first assign only real-world entities from the graph and passage, then determine the \(0/1\) interpretations after seeing the CPTs and the entity mapping. The final narrative-generation prompt remains the same as Prompt~2. After generation, automatic validators reject outputs with symbolic-token leakage or insufficient numeric fidelity, and failed records are retried before final dataset materialization.

\end{document}